\pdfoutput=1

\documentclass[11pt]{article}

\usepackage[final]{acl}

\usepackage{times}
\usepackage{latexsym}

\usepackage[T1]{fontenc}

\usepackage[utf8]{inputenc}

\usepackage{microtype}

\usepackage{inconsolata}

\usepackage{graphicx}

\usepackage{booktabs}
\usepackage{arydshln}
\usepackage{amssymb}
\usepackage[most]{tcolorbox}
\usepackage{colortbl}
\usepackage{subcaption}
\usepackage{dsfont}
\usepackage{multirow}
\usepackage{tabularx}

\usepackage{listings}
\usepackage{xcolor}
\colorlet{punct}{red!60!black}
\definecolor{background}{HTML}{EEEEEE}
\definecolor{delim}{RGB}{20,105,176}
\colorlet{numb}{magenta!60!black}
\lstdefinelanguage{json}{
    basicstyle=\normalfont\ttfamily,
    numbers=left,
    numberstyle=\scriptsize,
    stepnumber=1,
    numbersep=8pt,
    showstringspaces=false,
    breaklines=true,
    frame=lines,
    backgroundcolor=\color{background},
    literate=
     *{0}{{{\color{numb}0}}}{1}
      {1}{{{\color{numb}1}}}{1}
      {2}{{{\color{numb}2}}}{1}
      {3}{{{\color{numb}3}}}{1}
      {4}{{{\color{numb}4}}}{1}
      {5}{{{\color{numb}5}}}{1}
      {6}{{{\color{numb}6}}}{1}
      {7}{{{\color{numb}7}}}{1}
      {8}{{{\color{numb}8}}}{1}
      {9}{{{\color{numb}9}}}{1}
      {:}{{{\color{punct}{:}}}}{1}
      {,}{{{\color{punct}{,}}}}{1}
      {\{}{{{\color{delim}{\{}}}}{1}
      {\}}{{{\color{delim}{\}}}}}{1}
      {[}{{{\color{delim}{[}}}}{1}
      {]}{{{\color{delim}{]}}}}{1},
}

\newcommand\setrow[1]{\gdef\rowmac{#1}#1\ignorespaces}
\newcommand\clearrow{\global\let\rowmac\relax}
\clearrow

\definecolor{in}{HTML}{636EFA}
\definecolor{ian}{HTML}{FF6692}
\definecolor{rp}{HTML}{FECB52}
\definecolor{ip}{HTML}{FFA15A}
\definecolor{ianp}{HTML}{EF553B}
\definecolor{inp}{HTML}{AB63FA}
\definecolor{dgreen}{HTML}{009B55}

\definecolor{user}{HTML}{b3e5fc}
\definecolor{sys}{HTML}{616161}

\newcommand{\error}[1]{\textcolor{red}{#1}}

\newtcbox{\dau}[1][]{enhanced,
 box align=base,
 nobeforeafter,
 colback=user,
 colframe=black,
 size=small,
 fontupper=\ttfamily,
 left=0pt,
 right=0pt,
 boxsep=3pt,
 #1}

\newtcbox{\das}[1][]{enhanced,
 box align=base,
 nobeforeafter,
 colback=white,
 colframe=black,
 size=small,
 fontupper=\ttfamily,
 left=0pt,
 right=0pt,
 boxsep=3pt,
 #1}

\title{Dialog2Flow: Pre-training Soft-Contrastive Action-Driven Sentence Embeddings for Automatic Dialog Flow Extraction}

\author{Sergio Burdisso$^{1}$, Srikanth Madikeri$^{1,2}$ \and Petr Motlicek$^{1,3}$ \\
         $^1$Idiap Research Institute, Martigny, Switzerland \\ $^2$Department of Computational Linguistics, University of Zurich, Zurich, Switzerland\\$^3$Brno University of Technology, Brno, Czech Republic\\ 
         \texttt{sergio.burdisso@idiap.ch}}

\begin{document}
\maketitle

\begin{abstract}
Efficiently deriving structured workflows from unannotated dialogs remains an underexplored and formidable challenge in computational linguistics. Automating this process could significantly accelerate the manual design of workflows in new domains and enable the grounding of large language models in domain-specific flowcharts, enhancing transparency and controllability.
In this paper, we introduce Dialog2Flow (D2F) embeddings, which differ from conventional sentence embeddings by mapping utterances to a latent space where they are grouped according to their communicative and informative functions (i.e., the actions they represent). D2F allows for modeling dialogs as continuous trajectories in a latent space with distinct action-related regions. By clustering D2F embeddings, the latent space is quantized, and dialogs can be converted into sequences of region/action IDs, facilitating the extraction of the underlying workflow.
To pre-train D2F, we build a comprehensive dataset by unifying twenty task-oriented dialog datasets with normalized per-turn action annotations. We also introduce a novel soft contrastive loss that leverages the semantic information of these actions to guide the representation learning process, showing superior performance compared to standard supervised contrastive loss.
Evaluation against various sentence embeddings, including dialog-specific ones, demonstrates that D2F yields superior qualitative and quantitative results across diverse domains.\footnote{\url{https://github.com/idiap/dialog2flow}}
\end{abstract}

\section{Introduction}

Conversational AI has seen significant advancements, especially with the rise of Large Language Models (LLMs)~\cite{bubeck2023sparks,lu2022learn,hendryckstest2021,hendrycksmath2021,cobbe2021training}. Dialog modeling can be divided into open-domain dialogs and task-oriented dialogs (TOD), with the latter focusing on helping users achieve specific tasks~\cite{jurafsky2006pragmatics}.
In TOD, structured workflows guide agents in assisting users effectively.
This paper explores the underexplored terrain of automatically extracting such workflow from a collection of conversations.

\begin{figure}[t]
        \centering
        \small
        \begin{tabular}{p{.95\linewidth}}
            \toprule
            \rowcolor{user}
            \textbf{User:} i'm looking for the transplant unit department please \\
            \rowcolor{user}
           {\tiny \hspace{0.1in} \textbf{Action:} \dau{\textbf{INFORM} DEPARTMENT}} \\
            \textbf{System:} okay the transfer unit department give me a second let me look okay yes i found the transplant unit department can i help \\
            {\tiny \hspace{0.1in} \textbf{Action:} \das{\textbf{REQMORE}}} \\
        
            \rowcolor{user}
            \textbf{User:} may you please provide me with the phone number please \\
            \rowcolor{user}
           {\tiny \hspace{0.1in} \textbf{Action:} \dau{\textbf{REQUEST} PHONE}} \\
            \textbf{System:} get no problem okay so the number is 1223217711 \\
            {\tiny \hspace{0.1in} \textbf{Action:} \das{\textbf{INFORM} PHONE}} \\
        
            \rowcolor{user}
            \textbf{User:} okay um just repeat it it's 1 2 2 3 2 1 7 1 1 \\
            \rowcolor{user}
           {\tiny \hspace{0.1in} \textbf{Action:} \dau{\textbf{CONFIRM} PHONE}} \\
            \textbf{System:} okay thank you very much \\
            {\tiny \hspace{0.1in} \textbf{Action:} \das{\textbf{THANK\_YOU}}} \\
            \bottomrule
        \end{tabular}
        \caption{Example segment of the dialog \texttt{SNG1533} from the \texttt{hospital} domain of the SpokenWOZ dataset. Actions are defined by concatenating the dialog act label (in bold) with the slot label(s) associated to each utterance.}
\label{fig:dialog-example}
\end{figure}

Extracting workflows automatically is crucial for enhancing dialog system design, discourse analysis, data augmentation~\cite{qiu2022structure}, and training human agents~\cite{sohn-etal-2023-tod}. Additionally, it can ground LLMs in domain-specific workflows, improving transparency and control~\cite{raghu-etal-2021-end,chen2024benchmarking}.
Recent works have attempted to induce structural representations from dialogs using either ground truth annotation or \textit{ad hoc} methods~\cite{wd2023,qiu2022structure,sun2021unsupervised,qiu-etal-2020-structured}.
We believe that models specifically pre-trained for this purpose could significantly advance the field.

\begin{figure}[t]
    \centering
    \includegraphics[width=.95\linewidth]{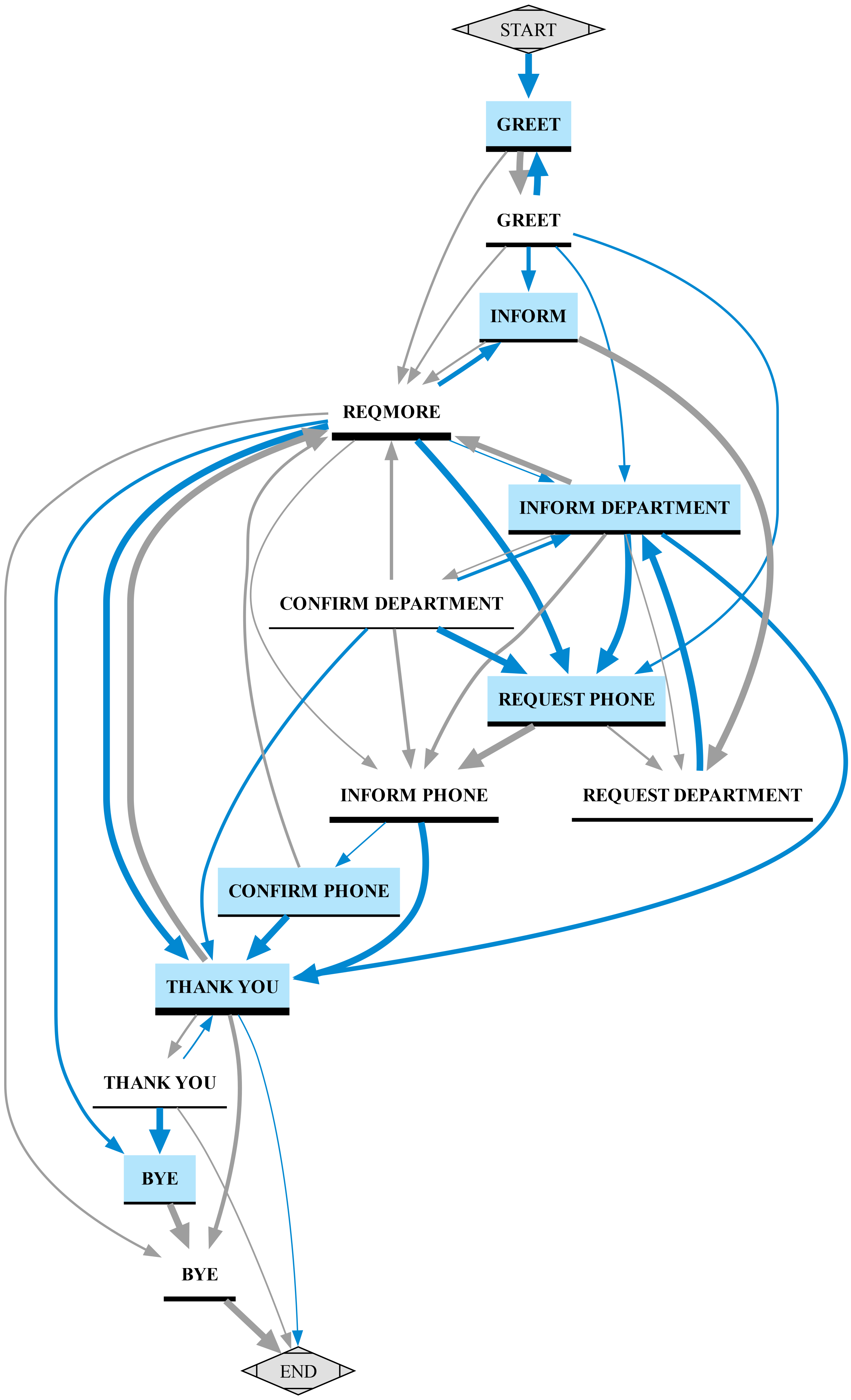}
    \caption{Directed graph representing the \texttt{hospital} domain workflow obtained from all the \texttt{hospital} dialogs in the SpokenWOZ dataset. Nodes correspond to individual actions. The width of edges and the underline thickness of nodes indicate their frequency. User actions are colored to distinguish them from system actions.}
    \label{fig:gt-graph}
\end{figure}

In Task-Oriented Dialog (TOD), \textit{dialog acts} and \textit{slots} are key concepts \cite{jurafsky2006pragmatics}.
Dialog acts represent the speaker's communicative intent, while slots capture task-specific information.
A \textit{dialog action} encapsulates both the dialog act and its corresponding slots, enabling us to view dialogs as sequences of canonical \textit{steps} that convey both \textit{communicative and informative functions} (Figure~\ref{fig:dialog-example}).
Motivated by this perspective, we propose embedding sentences into a latent space grouped by representative \textit{actions} rather than solely by sentence semantics.
Similar to how aggregating action sequences from multiple dialogs reveals a common underlying workflow (Figure~\ref{fig:gt-graph}), clustering sentence embeddings in this latent space could uncover common conversational steps, potentially revealing the underlying workflow.
The main contributions of this work are threefold: (a) we consolidate twenty task-oriented dialog datasets to create the largest publicly available dataset with standardized action annotations; (b) we introduce a novel \textit{soft} contrastive loss that leverages the semantic information of dialog actions to guide the representation learning process, outperforming standard supervised contrastive loss; and (c) we introduce and release Dialog2Flow (D2F), to the best of our knowledge, the first sentence embedding model pre-trained specifically for dialog flow extraction.

\section{Related Work}



\noindent
\textbf{Sentence Embeddings}~~Transformer-based encoders like Universal Sentence Encoder~\cite{cer2018universal} and Sentence-BERT~\cite{reimers-gurevych-2019-sentence} outperformed RNN-based ones such as Skip-Thought~\cite{kiros2015skip} and InferSent~\cite{conneau2017supervised}.
These models use a \textit{pooling strategy} (e.g., mean pooling, \texttt{[CLS]} token) to obtain a single sentence embedding optimized for semantic similarity. However, specific domains require different similarity notions.
In the context of dialogs, models like TOD-BERT~\cite{wu-etal-2020-tod}, DialogueCSE~\cite{dialoguecse2021} and Dialog Sentence Embedding (DSE)~\cite{zhou-etal-2022-learning} have shown that conversation-based similarity outperforms semantic similarity across different TOD tasks.
Likewise, we hypothesize that action-based similarity can yield meaningful workflow-related sentence embeddings.


\noindent
\textbf{Contrastive Learning}~~Contrastive learning has achieved success in representation learning for both images~\cite{chen2020simple,He2020moco,henaff2020data,tian2020contrastive,chen2020simple,hjelm2018learning} and text~\cite{zhou-etal-2022-learning,zhang-etal-2022-virtual,zhang-etal-2021-pairwise,gao-etal-2021-simcse,wu-etal-2020-tod}. It learns a representation space where similar instances cluster together and dissimilar instances are separated. More precisely, given an \textit{anchor} with \textit{positive} and \textit{negative} counterparts, the goal is to minimize the distance between anchor-positive pairs while maximizing the distance between anchor-negative pairs. Negatives are typically obtained through in-batch negative sampling, where positives from different anchors in the mini-batch are used as negatives.

\section{Method}





\subsection{Representation Learning Framework}

Following common practices~\cite{zhou-etal-2022-learning,chen2020simple,tian2020contrastive,khosla2020supervised}, the main components of our framework are:

\noindent
$\bullet$~\textbf{Encoder}, $f(\cdot)\in \mathds{R}^n$, which maps $x$ to a representation vector, $\mathbf{x} = f(x)$.
Following Sentence-BERT~\cite{reimers-gurevych-2019-sentence} and DSE~\cite{reimers-gurevych-2019-sentence}, $f(\cdot)$ consists of a BERT-based encoder with mean pooling strategy trained as a bi-encoder with shared weights (siamese network). 

\noindent
$\bullet$~\textbf{Contrastive head}, $g(\cdot)\in \mathds{R}^d$, used during training to map representations $\mathbf{x}$ to the space where contrastive loss is applied.
Following \citet{chen2020simple} and DSE, we instantiate $g(\cdot)$ as the multi-layer perceptron with a single hidden layer $\mathbf{z} = g(\mathbf{x}) = \text{ReLU}(\mathbf{x}\cdot W_1) W_2$ where $W_1 \in \mathds{R}^{n \times n}$ and $W_2 \in \mathds{R}^{n \times d}$.

\noindent
$\bullet$~\textbf{Similarity measure}, $\textit{sim}(\mathbf{u},\mathbf{v})$, used to learn the representation is cosine similarity. Thus, similarity is then measured only by the angle between $\mathbf{u}$ and $\mathbf{v}$, making our latent space geometrically a unit hypersphere. Hence, in this study, we treat similarity and alignment interchangeably. Additionally, we assume $f(\cdot)$ and $g(\cdot)$ vectors are L2-normalized, leading to $\textit{sim}(\mathbf{u},\mathbf{v})=\textit{cos}(\mathbf{u},\mathbf{v})=\mathbf{u}\cdot\mathbf{v}$.

\subsubsection{Supervised Contrastive Loss}

For a batch of $N$ randomly sampled anchor, positive, and label triples, $B=\{(x_i, x_i^+, y_i)\}_{i=1}^N$, the supervised contrastive loss~\cite{khosla2020supervised}, for each $i$-th triplet $(x_i, x_i^+, y_i)$ is defined as:
\vspace{-.4\baselineskip}
\begin{equation}
    \ell_i^{sup} = -\sum_{j\in\mathcal{P}_i} \frac{1}{|\mathcal{P}_i|} \text{log} \frac{e^{\mathbf{z}_i \cdot \mathbf{z}_j^+ / \tau}}{\sum_{k=1}^N{e^{\mathbf{z}_i \cdot \mathbf{z}_k^+ / \tau}}}
\label{eq:loss}
\end{equation}
\noindent
where $\mathcal{P}_i = \{j\mid y_i = y_j\}$ is the set of indexes of all the samples with the same label as the $i$-th sample in the batch, and $\tau$ is the softmax temperature parameter that controls how soft/strongly positive pairs are pulled together and negative pairs pushed apart in the embedding space.\footnote{The lower $\tau$, the sharper the softmax output distribution and the stronger the push/pull factor.}
The final loss is computed across all the $N$ pairs in the mini-batch as $\mathcal{L}^{sup} = \frac{1}{N}\sum_{i=1}^N \ell_i^{sup}$.

\subsubsection{Supervised \textit{Soft} Contrastive Loss}

Let $\delta(y_i, y_j)$ be a semantic similarity measure between labels $y_i$ and $y_j$. We define our soft contrastive loss as follows:
\begin{align*}
    \ell_i^{soft}\!=\!-\sum_{j=1}^N \frac{e^{\delta(y_i, y_j) / \tau'}}{\sum_{k=1}^N{e^{\frac{\delta(y_i, y_k)}{\tau'}}}} \text{log} \frac{e^{\mathbf{z}_i \cdot \mathbf{z}_j^+ / \tau}}{\sum_{k=1}^N{e^{\frac{\mathbf{z}_i \cdot \mathbf{z}_k^+}{\tau}}}}
\end{align*}
\noindent
where $\tau'$ is a temperature parameter controlling the "softness" of the negative labels (impact analysis available in Appendix~\ref{app:label_temperature}). 
For further details, including the underlying intuition behind the equation, please refer to Appendix~\ref{app:loss}. Unlike Equation~\ref{eq:loss}, this loss encourages the encoder to separate anchors and negatives \textit{proportionally to the semantic similarity of their labels}.
Finally, the mini-batch loss $\mathcal{L}^{soft}$ is computed as in $\mathcal{L}^{sup}$.

\subsection{Training Targets}

We experiment with four types of training targets, distinguished by whether the dialogue action label is used directly or decomposed into dialogue act and slot labels, and by the type of contrastive loss employed. Specifically, we consider the following two targets using the proposed soft contrastive loss:

\noindent
$\bullet$~\textbf{D2F$_{single}$}: $\mathcal{L} = \mathcal{L}^{soft}_{\textbf{act+slots}}$

\noindent
$\bullet$~\textbf{D2F$_{joint}$}: $\mathcal{L} = \mathcal{L}^{soft}_{\textbf{act}} + \mathcal{L}^{soft}_{\textbf{slots}}$

\noindent
and the two corresponding targets using the default supervised contrastive loss:

\noindent
$\bullet$~\textbf{D2F-Hard$_{single}$}: $\mathcal{L} = \mathcal{L}^{sup}_{\textbf{act+slots}}$

\noindent
$\bullet$~\textbf{D2F-Hard$_{joint}$}: $\mathcal{L} = \mathcal{L}^{sup}_{\textbf{act}} + \mathcal{L}^{sup}_{\textbf{slots}}$

\noindent
The subscript in bold indicates the type of label used to compute the loss, either the dialog action as a single label (\textit{act+slots}), or the dialog act and slots separately. In the case of the joint loss, separate contrastive heads $g(\cdot)$ are employed.

\section{Training Corpus}

\begin{table}[!t]
    \centering
    \small
    \begin{tabular}{l@{}c@{~}c@{~}c@{}c@{}}
        \toprule
        \textbf{Dataset} & \textbf{\#U} & \textbf{\#D} & \textbf{\#DA} & \textbf{\#S} \\
        \midrule
        ABCD~\cite{chen-etal-2021-action} & 20.4K & 10 & 0 & 10 \\
        BiTOD~\cite{bitod2021} & 72.5K & 6 & 13 & 33 \\
        Disambiguation~\cite{qian-etal-2022-database} & 114.3K & 8 & 9 & 28 \\
        DSTC2-Clean~\cite{mrksic-etal-2017-neural} & 25K & 1 & 2 & 8 \\
        FRAMES~\cite{el-asri-etal-2017-frames} & 20K & 1 & 21 & 46 \\
        GECOR~\cite{quan-etal-2019-gecor} & 2.5K & 1 & 2 & 10 \\
        HDSA-Dialog~\cite{chen-etal-2019-semantically} & 91.9K & 8 & 6 & 24 \\
        KETOD~\cite{chen-etal-2022-ketod} & 107.7K & 20 & 15 & 182 \\
        MS-DC~\cite{li2018microsoft} & 71.9K & 3 & 11 & 56 \\
        MulDoGO~\cite{peskov-etal-2019-multi} & 74.8K & 6 & 0 & 63 \\
        MultiWOZ2.1~\cite{eric-etal-2020-multiwoz} & 108.3K & 8 & 9 & 27 \\
        MultiWOZ2.2~\cite{zang-etal-2020-multiwoz} & 55.9K & 8 & 2 & 26 \\
        SGD~\cite{rastogi2020towards} & 479.5K & 20 & 15 & 184 \\
        Taskmaster1~\cite{byrne-etal-2019-taskmaster} & 30.7K & 6 & 1 & 59 \\
        Taskmaster2~\cite{byrne-etal-2019-taskmaster} & 147K & 11 & 1 & 117 \\
        Taskmaster3~\cite{byrne-etal-2019-taskmaster} & 589.7K & 1 & 1 & 21 \\
        WOZ2.0~\cite{mrksic-etal-2017-neural} & 4.4K & 1 & 2 & 10 \\
        SimJointMovie~\cite{shah-etal-2018-bootstrapping} & 7.2K & 1 & 14 & 5 \\
        SimJointRestaurant~\cite{shah-etal-2018-bootstrapping} & 20K & 1 & 15 & 9 \\
        SimJointGEN~\cite{zhang-etal-2024-dialogstudio} & 1.3M & 1 & 16 & 5 \\

        \midrule
        \textbf{Total} & \textbf{3.4M} & \textbf{52} & \textbf{44} & \textbf{524} \\
        \bottomrule
    \end{tabular}
    \caption{Details of used TOD datasets, including the number of utterances (\#U), unique domains (\#D), dialog act labels (\#DA), and slot labels (\#S).}
    \label{tab:datasets}
\end{table}

We identified and collected 20 TOD datasets from which we could extract dialog act and/or slot annotations, as summarized in Table~\ref{tab:datasets}.
We then manually inspected each dataset to locate and extract the necessary annotations, manually standardizing domain names and dialog act labels across datasets.
Finally, we unified all datasets under a consistent format, incorporating per-turn dialog act and slot annotations. The resulting unified TOD dataset comprises 3.4 million utterances annotated with 18 standardized dialog acts, 524 unique slot labels, and 3,982 unique action labels (dialog act + slots) spanning across 52 different domains (details in Appendix~\ref{app:dataset}).


\section{Experimental Setup}

For training D2F we mostly follow the experimental setup of DSE~\cite{zhou-etal-2022-learning} and TOD-BERT~\cite{wu-etal-2020-tod}, using BERT$_{base}$ as the backbone model for the encoder to report results in the main text.
Additional configurations are reported in the ablation study (Appendix~\ref{app:ablation}) while implementation details are given in Appendix~\ref{app:hyperparameters}.

\subsection{Baselines}

\noindent
\textbf{General sentence embeddings.} 
$\bullet$~\textbf{\texttt{GloVe}}: the average of GloVe embeddings~\cite{pennington-etal-2014-glove}.
$\bullet$~\textbf{\texttt{BERT}}: the vanilla BERT$_{base}$ model with mean pooling strategy, corresponding to our untrained encoder.
$\bullet$~\textbf{\texttt{Sentence-BERT}}: the model with the best average performance reported among all Sentence-BERT pre-trained models, namely the \href{https://huggingface.co/sentence-transformers/all-mpnet-base-v2}{\texttt{all-mpnet-base-v2}} model pre-trained using MPNet~\cite{mpnet2020} and further fine-tuned on a 1 billion sentence pairs dataset.
$\bullet$~\textbf{\texttt{GTR-T5}}: the Generalizable T5-based dense Retriever~\cite{ni-etal-2022-large} pre-trained on a 2 billion web question-answer pairs dataset, outperforming previous sparse and dense retrievers on the BEIR benchmark~\cite{thakur2021beir}.
$\bullet$~\textbf{\texttt{OpenAI}}: the recently released \mbox{OpenAI's} \href{https://openai.com/blog/new-embedding-models-and-api-updates/}{\texttt{text-embedding-3-large}} model~\cite{openai2022text-online, openai2022text}


\noindent
\textbf{Dialog sentence embeddings.} 
$\bullet$~\textbf{\texttt{TOD-BERT}}: the \texttt{TOD-BERT-jnt} model reported in \citet{wu-etal-2020-tod} pre-trained to optimize a contrastive response selection objective by treating utterances and their dialog context as positive pairs.
The pre-training data is the combination of 9 publicly available task-oriented datasets around 1.4 million total utterances across 60 domains.
$\bullet$~\textbf{\texttt{DSE}}: pre-trained on the same dataset as TOD-BERT, DSE learns sentence embeddings by simply taking consecutive utterances of the same dialog as positive pairs for contrastive learning. DSE has shown to achieve better representation capability than the other dialog and general sentence embeddings on TOD downstream tasks~\cite{gung-etal-2023-intent,zhou-etal-2022-learning}.
$\bullet$~\textbf{\texttt{SBD-BERT}}: the \texttt{TOD-BERT-SBD$_{MWOZ}$} model reported in \citet{qiu2022structure} in which sentences are represented as the mean pooling of the tokens that are part of the slots of the utterance, as identified by a Slot Boundary Detection (SBD) model trained on the original MultiWOZ dataset~\cite{budzianowski-etal-2018-multiwoz}.
$\bullet$~\textbf{\texttt{DialogGPT}}: following TOD-BERT and DSE, we also report results with DialogGPT~\cite{zhang-etal-2020-dialogpt} using the mean pooling of its hidden states as the sentence representation.
$\bullet$~\textbf{\texttt{SPACE-2}}: a dialog representation model pre-trained on a corpus of 22.8 million utterances, 3.3 million of which are annotated with TOD labels~\cite{space2}. The annotation is used for supervised contrastive learning and follows a four-layer \textit{domain$\rightarrow$intent$\rightarrow$slot$\rightarrow$value} semantic tree structure.\footnote{Except \texttt{DSE} and \texttt{SBD-BERT}, models are optimized for dialog context and may underperform on isolated sentences due to reliance on dialog-specific features like turns and roles.}



\subsection{Evaluation Data}

Most of the TOD datasets are constructed solely based on written texts, which may not accurately reflect the nuances of real-world spoken conversations, potentially leading to a gap between academic research and real-world spoken TOD scenarios.
Therefore, we evaluate our performance not only on a subset of our unified TOD dataset but also on SpokenWOZ~\cite{si2023spokenwoz}, the first large-scale human-to-human speech-text dataset for TOD designed to address this limitation.
More precisely, we use the following two evaluation sets:

\noindent
$\bullet$~\textbf{Unified TOD evaluation set:} 26,910 utterances with 1,794 unique \textit{action} labels (dialog act + slots) extracted from the training data.
These utterances were extracted by sampling and removing 15 utterances for each \textit{action} label with more than 100 utterances in the training data.

\noindent
$\bullet$~\textbf{SpokenWOZ:} 31,303 utterances with 427 unique \textit{action} labels corresponding to all the 1,710 single domain conversations in SpokenWOZ.
We are only using complete single-domain conversations so that we can also use them later to extract the domain-specific workflow for each of the 7 domains in SpokenWOZ.\footnote{There are no single-domain calls for the \texttt{profile} domain so it is not included.}
%

\section{Similarity-based Evaluation}
\label{sec:evaluation-similarity}

Before the dialog flow-based evaluation, we assess the quality of the representation space geometry through the similarity of the embeddings representing different \textit{actions}. We use the following methods as quality proxies:


\noindent
\noindent
$\bullet$~\textbf{Anisotropy.} Following \citet{jiang-etal-2022-promptbert,ethayarajh-2019-contextual}, we measure the anisotropy of a set of embeddings as the average cosine (absolute) similarity among all embeddings in the set.\footnote{$\frac{1}{n^2-n}\left|\sum_i\sum_{j\neq i} \text{cos}(\mathbf{x}_i, \mathbf{x}_j) \right|$ for given $\{\mathbf{x}_1,\cdots,\mathbf{x}_n\}$} Ideally, embeddings of the same \textit{action} should be similar (high intra-action anisotropy) while being dissimilar to those of other actions (low inter-action anisotropy). We report the average intra- and inter-action anisotropy across all actions.

\noindent
$\bullet$~\textbf{Similarity-based few-shot classification.} 
We use Prototypical Networks~\cite{snell2017prototypical} to perform a similarity-based classification. A prototype embedding for each \textit{action} is calculated by averaging $k$ of its embeddings ($k$-shot). All other embeddings are then classified based on the closest prototype embedding. We report the \textit{macro averaged F}$_1$ score and \textit{Accuracy} for $k=1$ and $k=5$ (i.e., 1-shot and 5-shot classification).


\noindent
$\bullet$~\textbf{Ranking.} For each action, we randomly select one utterance as the query and retrieve the top-$k$ closest embeddings, creating a ranking with their actions. Ideally, the top-$k$ retrieved embeddings should predominantly correspond to the same \textit{action} as the query, thus ranked first. We report \textit{Normalized Discounted Cumulative Gain} (nDCG@10), averaged over all actions.

\subsection{Similarity-based Results}

\begin{table*}[!t]
    \small
    \centering
    \begin{tabular}{lcc|cc|ccc}
        \toprule
        & \multicolumn{2}{c}{\textbf{F$_1$ score}} & \multicolumn{2}{c}{\textbf{Accuracy}} & \multicolumn{3}{c}{\textbf{Anisotropy}} \\
        \textbf{Embeddings} & \textit{1-shot} & \textit{5-shot} & \textit{1-shot} & \textit{5-shot} & \textit{intra}($\uparrow$) & \textit{inter}($\downarrow$) & $\Delta$ ($\uparrow$)  \\
        \midrule
        GloVe & 23.24 $\pm$ 0.87 & 24.45 $\pm$ 0.94 & 26.04 $\pm$ 0.81 & 30.01 $\pm$ 0.86 & 0.674 & 0.633 & 0.041 \\
        BERT & 23.85 $\pm$ 0.47 & 28.22 $\pm$ 0.60 & 26.32 $\pm$ 0.62 & 32.92 $\pm$ 0.38 & 0.737 & 0.781 & -0.044 \\
        Sentence-BERT & 27.86 $\pm$ 0.93 & 33.30 $\pm$ 0.68 & 30.55 $\pm$ 0.82 & 38.22 $\pm$ 0.46 & 0.527 & \textbf{0.433} & 0.094 \\
        GTR-T5 & 30.86 $\pm$ 0.39 & 38.38 $\pm$ 0.64 & 33.34 $\pm$ 0.29 & 42.96 $\pm$ 0.60 & 0.694 & 0.706 & -0.012 \\
        OpenAI & 32.12 $\pm$ 0.87 & 41.06 $\pm$ 0.68 & 34.95 $\pm$ 0.84 & 45.51 $\pm$ 0.60 & 0.541 & 0.424 & \textbf{0.117} \\
        DSE & \textbf{35.43} $\pm$ 0.96 & \textbf{42.21} $\pm$ 0.90 & \textbf{38.12} $\pm$ 0.77 & \textbf{46.85} $\pm$ 0.79 & 0.649 & 0.541 & 0.108 \\
        SPACE-2 & 26.93 $\pm$ 0.64 & 37.04 $\pm$ 0.66 & 28.95 $\pm$ 0.62 & 41.32 $\pm$ 0.57 & 0.664 & 0.646 & 0.018 \\
        TOD-BERT & 27.58 $\pm$ 0.92 & 33.35 $\pm$ 0.58 & 29.63 $\pm$ 1.06 & 36.88 $\pm$ 0.87 & \textbf{0.840} & 0.864 & -0.024 \\
        DialoGPT & 25.86 $\pm$ 0.34 & 31.34 $\pm$ 0.73 & 28.24 $\pm$ 0.53 & 36.15 $\pm$ 0.83 & 0.734 & 0.758 & -0.024 \\
        SBD-BERT & 24.31 $\pm$ 0.95 & 27.71 $\pm$ 0.38 & 26.40 $\pm$ 0.96 & 31.53 $\pm$ 0.44 & 0.687 & 0.604 & 0.083 \\
        \midrule
        D2F-Hard$_{single}$ & \textbf{58.84} $\pm$ 0.62 & \textbf{67.82} $\pm$ 0.52 & \textbf{61.52} $\pm$ 0.54 & \textbf{70.69} $\pm$ 0.43 & \textbf{0.646} & \textbf{0.313} & \textbf{0.332} \\
        D2F-Hard$_{joint}$ & 56.25 $\pm$ 1.16 & 66.22 $\pm$ 0.62 & 58.98 $\pm$ 1.08 & 69.23 $\pm$ 0.48 & 0.629 & 0.399 & 0.230 \\
        \addlinespace
        \cdashline{2-8}
        \addlinespace
        D2F$_{single}$ & \underline{\textbf{65.36}} $\pm$ 0.91 & 70.89 $\pm$ 0.30 & \underline{\textbf{68.06}} $\pm$ 0.87 & \underline{\textbf{74.15}} $\pm$ 0.40 & \textbf{0.782} & \underline{\textbf{0.186}} & \underline{\textbf{0.597}} \\
        D2F$_{joint}$ & 63.70 $\pm$ 1.35 & \underline{\textbf{70.94}} $\pm$ 0.41 & 66.53 $\pm$ 1.15 & 74.03 $\pm$ 0.31 & 0.741 & 0.289 & 0.451 \\
        \bottomrule
    \end{tabular}
    \caption{Similarity-based few-shot classification results on our unified TOD evaluation set. The intra- and inter-action anisotropy are also provided along their difference ($\Delta$). \textbf{Bold} indicates the best values in each group while \underline{\textbf{underlined}} the global best.}
    \label{tab:results-in-domain}
\end{table*}

\begin{figure*}[t!]
    \centering
    \begin{subfigure}[t]{0.245\textwidth}
        \centering
        \includegraphics[height=1.0in]{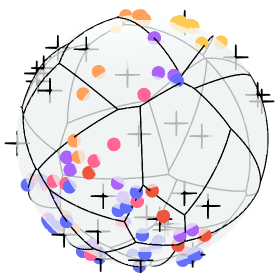}
        \caption{\textbf{Sentence-BERT}}
    \end{subfigure}%
    \begin{subfigure}[t]{0.245\textwidth}
        \centering
        \includegraphics[height=1.0in]{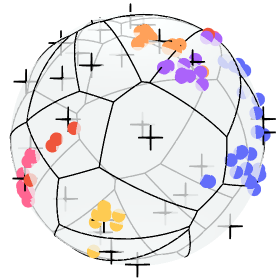}
        \caption{\textbf{D2F-Hard$_{joint}$}}
    \end{subfigure}
    \begin{subfigure}[t]{0.245\textwidth}
        \centering
        \includegraphics[height=1.0in]{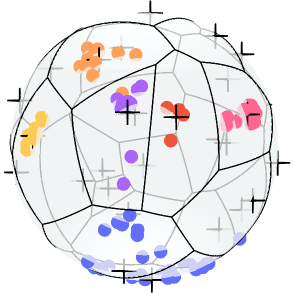}
        \caption{\textbf{D2F$_{joint}$}}
    \end{subfigure}
    \begin{subfigure}[t]{0.245\textwidth}
        \centering
        \includegraphics[height=1in]{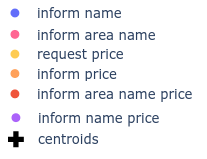}
    \end{subfigure}

    \caption{Spherical Voronoi diagram of embeddings projected onto the unit sphere using UMAP with cosine distance as the metric. The embeddings represent system utterances from the \texttt{hotel} domain of the MultiWOZ2.1 dataset. Legends indicate the ground-truth action associated to each embedding and the centroids used to generate the partitions for all the actions in this domain.}
    \label{fig:umap-plots}
\end{figure*}

\begin{table*}[!ht]
    \centering
    \small
    \begin{tabular}{lcc|cc|ccc}
        \toprule
        & \multicolumn{2}{c}{\textbf{F$_1$ score}} & \multicolumn{2}{c}{\textbf{Accuracy}} & \multicolumn{3}{c}{\textbf{Anisotropy}} \\
        \textbf{Embeddings} & \textit{1-shot} & \textit{5-shot} & \textit{1-shot} & \textit{5-shot} & \textit{intra}($\uparrow$) & \textit{inter}($\downarrow$) & $\Delta$ ($\uparrow$)  \\
        \midrule
        GloVe & 19.47 $\pm$ 2.47 & 24.54 $\pm$ 2.45 & 26.07 $\pm$ 4.52 & 33.30 $\pm$ 4.19 & 0.653 & 0.642 & 0.010 \\
        BERT & 21.93 $\pm$ 2.40 & 31.11 $\pm$ 2.56 & 28.33 $\pm$ 3.76 & 39.98 $\pm$ 3.56 & 0.711 & 0.761 & -0.049 \\
        Sentence-BERT & 23.48 $\pm$ 2.62 & 35.71 $\pm$ 2.94 & 33.03 $\pm$ 4.70 & 47.47 $\pm$ 3.60 & 0.440 & \textbf{0.404} & 0.036 \\
        GTR-T5 & 26.53 $\pm$ 2.29 & 41.10 $\pm$ 2.37 & 35.76 $\pm$ 4.00 & 52.73 $\pm$ 3.16 & 0.681 & 0.714 & -0.033 \\
        OpenAI & \textbf{28.67} $\pm$ 2.33 & \textbf{42.49} $\pm$ 2.54 & \textbf{39.98} $\pm$ 3.77 & \textbf{55.37} $\pm$ 3.24 & 0.496 & 0.468 & 0.029 \\        
        DSE & 27.53 $\pm$ 2.70 & 39.90 $\pm$ 3.08 & 35.93 $\pm$ 4.54 & 51.73 $\pm$ 3.41 & 0.633 & 0.608 & 0.026 \\
        SPACE-2 & 25.07 $\pm$ 2.06 & 38.31 $\pm$ 2.38 & 34.00 $\pm$ 3.91 & 48.45 $\pm$ 3.21 & 0.653 & 0.650 & 0.003 \\
        TOD-BERT & 21.23 $\pm$ 2.03 & 32.28 $\pm$ 2.33 & 29.26 $\pm$ 3.99 & 41.71 $\pm$ 3.68 & \textbf{0.848} & 0.885 & -0.038 \\
        DialoGPT & 21.74 $\pm$ 2.10 & 32.01 $\pm$ 2.38 & 27.65 $\pm$ 3.47 & 41.05 $\pm$ 3.64 & 0.700 & 0.726 & -0.026 \\
        SBD-BERT & 19.09 $\pm$ 2.10 & 23.83 $\pm$ 2.22 & 25.80 $\pm$ 3.56 & 32.14 $\pm$ 3.62 & 0.651 & 0.596 & \textbf{0.055} \\
        \midrule
        D2F-Hard$_{single}$ & \textbf{34.64} $\pm$ 2.90 & \textbf{49.63} $\pm$ 2.87 & \textbf{42.77} $\pm$ 4.61 & \textbf{58.63} $\pm$ 3.27 & \textbf{0.526} & \textbf{0.424} & \textbf{0.103} \\
        D2F-Hard$_{joint}$ & 31.46 $\pm$ 2.61 & 46.89 $\pm$ 2.50 & 39.45 $\pm$ 4.22 & 56.43 $\pm$ 2.98 & 0.514 & 0.481 & 0.033 \\
        \addlinespace
        \cdashline{2-8}
        \addlinespace
        D2F$_{single}$ & \underline{\textbf{35.55}} $\pm$ 3.51 & \underline{\textbf{49.75}} $\pm$ 2.48 & \underline{\textbf{43.15}} $\pm$ 5.24 & \underline{\textbf{59.93}} $\pm$ 3.06 & 0.516 & \underline{\textbf{0.321}} & \underline{\textbf{0.195}} \\
        D2F$_{joint}$ & 33.19 $\pm$ 2.95 & 46.90 $\pm$ 2.66 & 41.22 $\pm$ 4.40 & 57.07 $\pm$ 2.92 & \textbf{0.545} & 0.429 & 0.116 \\
        \bottomrule
    \end{tabular}
    \caption{Similarity-based few-shot classification results on SpokenWOZ. The intra- and inter-action anisotropy are also provided along their difference ($\Delta$).}
    \label{tab:results-out-domain}
\end{table*}

Tables~\ref{tab:results-in-domain} and \ref{tab:results-out-domain} present the similarity-based classification and anisotropy results on the unified TOD evaluation set and SpokenWOZ, respectively. Results are averaged over 1,794 and 427 different action labels for both datasets, respectively. For classification results, we report the mean and standard deviation from 10 repetitions, each sampling different embeddings for the 1-shot and 5-shot prototypes.
All D2F variants consistently outperform the baselines across all metrics.
This is expected, as D2F models, unlike the baselines, are explicitly trained to learn a representation space where embeddings are clustered by their corresponding actions.
However, the baseline results serve as a proxy for assessing the inherent suitability of existing sentence embedding models for our objective.\footnote{Throughout this paper, baseline results are intended to provide the reader with insights into the potential usability of available sentence embedding models if they were to be used for automatic dialog flow extraction, compared to our task-adaptive pre-trained embeddings~\cite{gururangan-etal-2020-dont}.}
For instance, as shown in Table~\ref{tab:results-in-domain}, DSE, which clusters sentences based on conversational context similarity (\textit{i.e.}, how often they appear consecutively in task-oriented dialogs), outperforms general-purpose embeddings that rely on semantic similarity.
Notably, D2F embeddings trained with the proposed soft contrastive loss exhibit superior performance compared to D2F-Hard embeddings trained with the standard supervised contrastive loss.
%
In Table~\ref{tab:results-out-domain}, the difference among the various embeddings narrows, and standard deviations increase significantly compared to Table~\ref{tab:results-in-domain}. This indicates that results vary considerably depending on the sampled prototypes, suggesting that the SpokenWOZ data is considerably noisier than the unified TOD evaluation set. This is expected as SpokenWOZ utterances were obtained by an ASR model from real-world human-to-human spoken TOD conversations, thus affected by ASR noise and various linguistic phenomena such as back-channels, disfluencies, and incomplete utterances.\footnote{Indeed, SpokenWOZ authors conducted experiments using newly proposed LLMs and dual-modal models, showing that current models face challenges on this more-realistic spoken dataset~\cite{si2023spokenwoz}.}

Classification results provide a local view of the representation space quality around the different sampled prototypes. Actions spread into multiple sub-clusters could still yield good classification results.
Thus, we also consider anisotropy results for a more global view of the representation space quality.
Among the baselines, TOD-BERT has the highest intra-action anisotropy but also the highest inter-action value, which means that, on average, embeddings of different actions are closer than embeddings of the same action! (negative $\Delta$ values). Sentence-BERT has the lowest inter-action anisotropy, indicating different actions are the most dissimilar, although embeddings of the same action are less similar ($\Delta = 0.094$) compared to DSE ($\Delta = 0.108$) in Table~\ref{tab:results-in-domain}. D2F embeddings exhibit the best anisotropy values, with a $\Delta$ difference between intra- and inter-action embeddings of $0.597$ and $0.451$ in Table~\ref{tab:results-in-domain}, and $0.193$ and $0.103$  in Table~\ref{tab:results-out-domain}, for single and joint targets, respectively, roughly doubling their D2F-Hard counterparts.

\

We hypothesize that the performance improvement observed when using the proposed soft-contrastive loss (\textit{i.e.}, D2F vs. D2F-Hard) stems from a more semantically informed arrangement of embeddings within the representation space. By leveraging action semantics during training, the soft-contrastive loss guides the learning process towards a more meaningful organization of embeddings.
For instance, Figure~\ref{fig:umap-plots} shows the projection of the embeddings onto the unit sphere for a subset of six related actions.\footnote{The original manifold in which utterances are embedded correspond to the unit hyper-sphere, thus, we believe the unit sphere provides a more truthful visualization than a 2D plane.} Sentence-BERT clusters embeddings into roughly two main semantic groups, with price-related actions on top and others at the bottom. D2F-Hard correctly clusters embeddings of the same action together while maintaining separation among centroids of different actions. However, the arrangement among different clusters is better in D2F, guided by action semantics --namely, all clusters are adjacent, with \textcolor{rp}{$\bullet$}[\texttt{request \textbf{price}}] next to \textcolor{ip}{$\bullet$}[\texttt{inform \textbf{price}}]; \textcolor{inp}{$\bullet$}[\texttt{inform \textbf{name price}}] between \textcolor{in}{$\bullet$}[\texttt{inform \textbf{name}}] and \textcolor{ip}{$\bullet$}[\texttt{inform \textbf{price}}]; and \textcolor{ianp}{$\bullet$}[\texttt{inform \textbf{name price area}}] between \textcolor{inp}{$\bullet$}[\texttt{inform \textbf{name price}}] and \textcolor{ian}{$\bullet$}[\texttt{inform \textbf{name area}}].

\begin{table}[!t]
    \centering
    \small
    \begin{tabular}{lcc}
        \toprule
        \textbf{Embeddings} & \textbf{NDCG@10}$^\clubsuit$ & \textbf{NDCG@10}$^\bigstar$ \\
        \midrule
        GloVe & 26.55 $\pm$ 0.57 & 25.09 $\pm$ 2.28 \\
        BERT & 26.98 $\pm$ 0.80 & 27.74 $\pm$ 2.00 \\
        Sentence-BERT & 30.88 $\pm$ 0.70 & 30.07 $\pm$ 2.23 \\
        GTR-T5 & 33.21 $\pm$ 0.60 & 32.74 $\pm$ 2.44 \\
        OpenAI & 35.82 $\pm$ 0.62 & \textbf{34.52} $\pm$ 2.01 \\
        DSE & \textbf{38.09} $\pm$ 0.71 & 33.94 $\pm$ 2.47 \\
        SPACE-2 & 30.01 $\pm$ 0.48 & 30.58 $\pm$ 2.01 \\
        TOD-BERT & 30.55 $\pm$ 0.74 & 25.63 $\pm$ 1.88 \\
        DialoGPT & 28.86 $\pm$ 0.71 & 27.92 $\pm$ 2.01 \\
        SBD-BERT & 27.20 $\pm$ 0.83 & 22.24 $\pm$ 1.93 \\
        \midrule
        D2F-Hard$_{single}$ & \textbf{60.87} $\pm$ 0.47 & \textbf{42.48} $\pm$ 2.77 \\
        D2F-Hard$_{joint}$ & 58.38 $\pm$ 0.72 & 40.03 $\pm$ 2.52 \\
        \addlinespace
        \cdashline{2-3}
        \addlinespace
        D2F$_{single}$ & \underline{\textbf{67.31}} $\pm$ 0.42 & \underline{\textbf{43.12}} $\pm$ 2.92 \\
        D2F$_{joint}$ & 66.50 $\pm$ 0.49 & 40.97 $\pm$ 2.61 \\
        


        

        \bottomrule
    \end{tabular}
    \caption{Ranking-based results on the unified TOD evaluation set ($\clubsuit$) and SpokenWOZ ($\bigstar$).}
    \label{tab:results-ranking}
\end{table}

\begin{table*}[!t]
    \centering
    \tiny
    \begin{tabular}{r|p{0.27\linewidth}|p{0.3\linewidth}|p{0.28\linewidth}}
        \toprule
        \textbf{Rank} & \textbf{DSE} & \textbf{Sentence-BERT} & \textbf{D2F$_{single}$} \\
        \midrule
        1. & \error{-uh my phone number is 7 4$^\blacksquare$} & -okay may i have your phone number please\error{$^\square$}  & -please get their phone number\error{$^\square$} \\
        2. & -okay okay now please get your number  & -may i get your phone number  & -okay may i have your phone number please\error{$^\square$} \\
        3. & -okay may i have your phone number please\error{$^\square$}  & -okay may i know your telephone number please  & -okay may i know your telephone number please \\
        4. & -thank you on the phone number\error{$^\square$}   & \error{-okay can i please get your id number$^\clubsuit$}  & -may i get your phone number \\
        5. & -okay may i know your telephone number please   & \error{-okay may i have your phone name in case for cooking the table$^\bigstar$}  & -um can i please have their phone number\error{$^\square$} \\
        6. & -okay great emma please have your contact number   & -okay and may i have your number please  & -okay so may i have the phone number with me \\
        7. & \error{-my number is 2 10$^\blacksquare$}  & -okay and may i have your number please  & -okay i'm i also need phone number\error{$^\square$} \\
        8. & \error{-the number is you see$^\spadesuit$}  & -okay and may i have your number please  & -no problem um but for the information can i have your phone number \\
        9. & -okay and may i have your number please  & \error{-okay and your car number$^\heartsuit$}  & -thank you on the phone number\error{$^\square$} \\
        10. & -okay and may i have your number please  & -this product uh may i have your phone number please  & -okay can i get your phone number please to make that booking \\
        \bottomrule
    \end{tabular}
    \caption{Top-10 retrieved utterances on SpokenWOZ for the query \textit{"your phone please"} with action label [\texttt{request phone\_number}].
    Errors are highlighted in red with wrong action marked as:  $^\blacksquare$[\texttt{inform phone\_number}]; $^\spadesuit$[\texttt{inform plate\_number}]; $^\clubsuit$[\texttt{request id\_number}]; $^\bigstar$[\texttt{request name}]; $^\heartsuit$[\texttt{request plate\_number}]; $^\square$[\texttt{request phone}].}
    \label{tab:example-ranking}
\end{table*}

Finally, Table~\ref{tab:results-ranking} presents the ranking-based results on both evaluation sets. We report the mean and standard deviation from 10 repetitions, each sampling different query utterances for all actions. We observe a similar pattern across both datasets: an increase in variability and a drop in performance for all embedding types in SpokenWOZ. However, D2F embeddings still outperform all baselines and their D2F-Hard counterparts. For a more qualitative analysis, Table~\ref{tab:example-ranking} provides an example of the rankings obtained for the query \textit{"your phone please"} with the target action [\texttt{request phone\_number}] on SpokenWOZ. As seen, DSE errors arise due to embeddings being closer if they correspond to consecutive utterances (\texttt{inform} and \texttt{request} utterances). Sentence-BERT errors occur due to the retrieval of utterances semantically related to "number" and "phone." In contrast, all D2F-retrieved utterances correctly represent different ways to request a phone number, even though half were considered incorrect due to the lack of slot name standardization across different domains (e.g., \texttt{phone\_number} and \texttt{phone}).\footnote{Slot names mismatch across domains is also partially affecting all results reported in SpokenWOZ (Tables~\ref{tab:results-out-domain} and \ref{tab:results-ranking}).} Nonetheless, for clustering utterances by similarity to extract a dialog flow without annotation, D2F would successfully cluster these 10 utterances together as they correspond to semantically equivalent actions ([\texttt{request phone\_number}] and [\texttt{request phone}]).

\section{Dialog Flow Extraction Evaluation}
\label{sec:sec:evaluation-flow}

\begin{table*}[!t]
    \centering
    \small
    \begin{tabular}{l|c@{~~}c@{~~}c@{~~}c@{~}c@{~~}c|c}
    \toprule
    \textbf{Embeddings} & \textbf{Taxi} (31) & \textbf{Police} (23) & \textbf{Hospital} (18) & \textbf{Train} (49) & \textbf{Restaurant} (59) & \textbf{Attraction} (45) & \textbf{AVG.} \\
    \midrule
    D2F$_{single}$ & 9.68\%  (+3) & \underline{\textbf{4.35\%}} (-1) & \textbf{11.11\%} (-2) & \underline{\textbf{2.04\%}} (+1) & \underline{\textbf{5.08\%}} (-3) & \textbf{8.89\%} (+4) & \underline{\textbf{6.86\%}} \\
    D2F$_{joint}$ & \textbf{3.23\%} (+1) & \textbf{8.70\%} (-2) & \underline{\textbf{5.56\%}} (-1) & \textbf{10.20\%} (-5) & 23.73\% (-14) & \underline{\textbf{0.00\%}} (0) & \textbf{8.57\%} \\
    D2F-Hard$_{single}$ & 12.90\% (-4) & 26.09\% (-6) & 16.67\% (-3) & \textbf{10.20\%} (-5) & \textbf{10.17\%} (-6) & 15.56\% (+7) & 15.26\% \\
    D2F-Hard$_{joint}$ & \underline{\textbf{0.00\%}} (0) & \textbf{8.70\%} (-2) & 33.33\% (-6) & 20.41\% (-10) & 25.42\% (-15) & 13.33\% (-6) & 16.87\% \\
    \midrule
    DSE & 32.26\% (-10) & 17.39\% (-4) & 33.33\% (-6) & 30.61\% (-15) & 27.12\% (-16) & 26.67\% (-12) & 27.90\% \\
    SPACE-2 & 32.26\% (-10) & 30.43\% (-7) & 38.89\% (-7) & 18.37\% (-9) & 32.20\% (-19) & 33.33\% (-15) & 30.91\% \\
    DialoGPT & 32.26\% (-10) & 34.78\% (-8) & 22.22\% (-4) & 44.90\% (-22) & 64.41\% (-38) & 51.11\% (-23) & 41.61\% \\
    BERT & 54.84\% (-17) & 30.43\% (-7) & 22.22\% (-4) & 46.94\% (-23) & 59.32\% (-35) & 42.22\% (-19) & 42.66\% \\
    OpenAI & 54.84\% (-17) &  52.17\% (-12) & 55.56\% (-10) & 42.86\% (-21) & 49.15\% (-29) & 44.44\% (-20) & 49.84\%  \\
    Sentence-BERT & 48.39\% (-15) & 43.48\% (-10) & 55.56\% (-10) & 57.14\% (-28) & 50.85\% (-30) & 55.56\% (-25) & 51.83\% \\
    GTR-T5 & 41.94\% (-13) & 43.48\% (-10) & 66.67\% (-12) & 51.02\% (-25) & 61.02\% (-36) & 53.33\% (-24) & 52.91\% \\
    SBD-BERT & 77.42\% (-24) & 43.48\% (-10) & 38.89\% (-7) & 71.43\% (-35) & 86.44\% (-51) & 86.67\% (-39) & 67.39\% \\
    TOD-BERT & 74.19\% (-23) & 78.26\% (-18) & 55.56\% (-10) & 85.71\% (-42) & 83.05\% (-49) & 82.22\% (-37) & 76.50\% \\
    \bottomrule
    \end{tabular}
\caption{Comparison of induced graph size vs. reference graph size for each single-domain in SpokenWOZ, measured by the number of nodes (actions). The table shows the normalized absolute difference (\%) and raw difference in parentheses. Column headers indicate the size of each reference graph ($G_D$). Lower differences suggest a better match in graph complexity.}
\label{tab:graph-results}
\end{table*}

Dialog flow extraction is an underexplored hard-to-quantify and challenging task with nuances in definition.
However, to evaluate embedding quality, we formally define the problem as follows:
Let $\mathcal{U}$ and $\mathcal{A}$ denote sets of TOD utterances and actions, respectively.
Let $\mathcal{U}$ and $\mathcal{A}$ be sets of TOD utterances and actions, respectively.
Let $\alpha:\mathcal{U}\mapsto\mathcal{A}$ be a (usually unknown) function mapping an utterance to its corresponding action.
Let $d_i = (u_1,\cdots,u_k)$ be a dialog with $u_j \in \mathcal{U}$, and $t_i = (\alpha(u_1),\cdots,\alpha(u_k)) = (a_1,\cdots,a_k)$ its conversion to a sequence of actions, referred to as a \textit{trajectory}.
Given a set of $m$ dialogs, $D=\{d_1,\cdots,d_m\}$, and after conversion to a set of action trajectories, $D^t=\{t_1,\cdots,t_m\}$, the goal is to extract the common dialog flow by combining all the trajectories in $D^t$.
We represent the common dialog flow as a weighted actions transition graph~\cite{ferreira-2023-automatic}.\footnote{Even though having states as individual actions makes them non-Markovian, this graph is easy to interpret and directly links the quality of individual actions to the overall flow's quality.}
More precisely, the common flow is represented as a weighted graph $G_D = \langle A, E, w_A, w_E \rangle$ where $A$ is the set of actions, $E$ represents edges between actions, the edge weight $w_E(a_i, a_j) \in [0, 1]$ indicates how often $a_i$ is followed by $a_j$, and the action weight $w_A(a_i) \in [0, 1]$ is its normalized frequency.

\subsection{Evaluation Details}

For each domain in SpokenWOZ, we build and compare its reference graph $G_D$ against the induced graph $\hat{G}_D$ using different embeddings.
The reference graph $G_D$ is built from the trajectories $D^t$ generated using the ground truth action labels ---\textit{e.g.} Figure~\ref{fig:gt-graph} is indeed $G_{hospital}$.
In contrast, the induced graph $\hat{G}_D$ is built \textit{without any annotation} by clustering all the utterance embeddings in $D$ and using the cluster ids as action labels to generate the trajectories $\hat{D}^t$.
That is, for $G_D$, we have $\alpha(u_i)=a_i$, while for $\hat{G}_D$, we have $\alpha(u_i)=c_i$ where $c_i$ is the cluster id assigned to $\mathbf{u}_i$.
To compare the induced and reference graphs, we report the difference in the number of nodes between them as the evaluation metric.\footnote{One cluster id $c_i$ can correspond to multiple $a_i$s and vice versa, preventing a direct comparison between $\hat{G}_D$ and $G_D$.}
Despite its simplicity, this metric allows us to compare the complexity of the induced vs. reference graph in terms of their sizes (\textit{i.e.} the number of discovered/extracted actions by each embedding model). 
Furthermore, to avoid the influence of infrequently occurring utterances/actions on graph size, we prune them by removing all nodes $a$ with $w_A(a) < \epsilon = 0.02$ (noise threshold).

In practice, the total number of actions to cluster is unknown in advance.
For instance, a hierarchical clustering algorithm can be used to approximate this number (see Appendix~\ref{app:cluster}).
However, for evaluation purposes, we set the number of clusters in each domain to be equal to the ground truth number so that all the embeddings are evaluated under the same best-case scenario in which this number is known in advance.
Therefore, all the induced graphs are built and processed equally, making the input embeddings the only factor influencing the final graph.

\subsection{Dialog Flow Extraction Results}

Table~\ref{tab:graph-results} shows the results obtained when comparing the different extracted graphs.
We can see that graphs obtained with available sentence embedding models tend to underestimate the complexity of each domain, producing less meaningful graphs with fewer states/actions than their references.
We hypothesize this is due to available models grouping the utterances either by conversational
context or semantic similarity, thus, only allowing us to discover either semantic or conversational-context "steps" (clusters/actions) in the dialogs from each domain.
For instance, Figure~\ref{fig:sbert-graph} and \ref{fig:dse-graph} in Appendix show the extracted graphs $\hat{G}_{hospital}$ with Sentence-BERT and DSE containing 10 and 6 less nodes ("steps") than the reference graph (Figure~\ref{fig:gt-graph}), respectively.

\begin{figure}[!ht]
    \centering
    \includegraphics[width=1.1\linewidth]{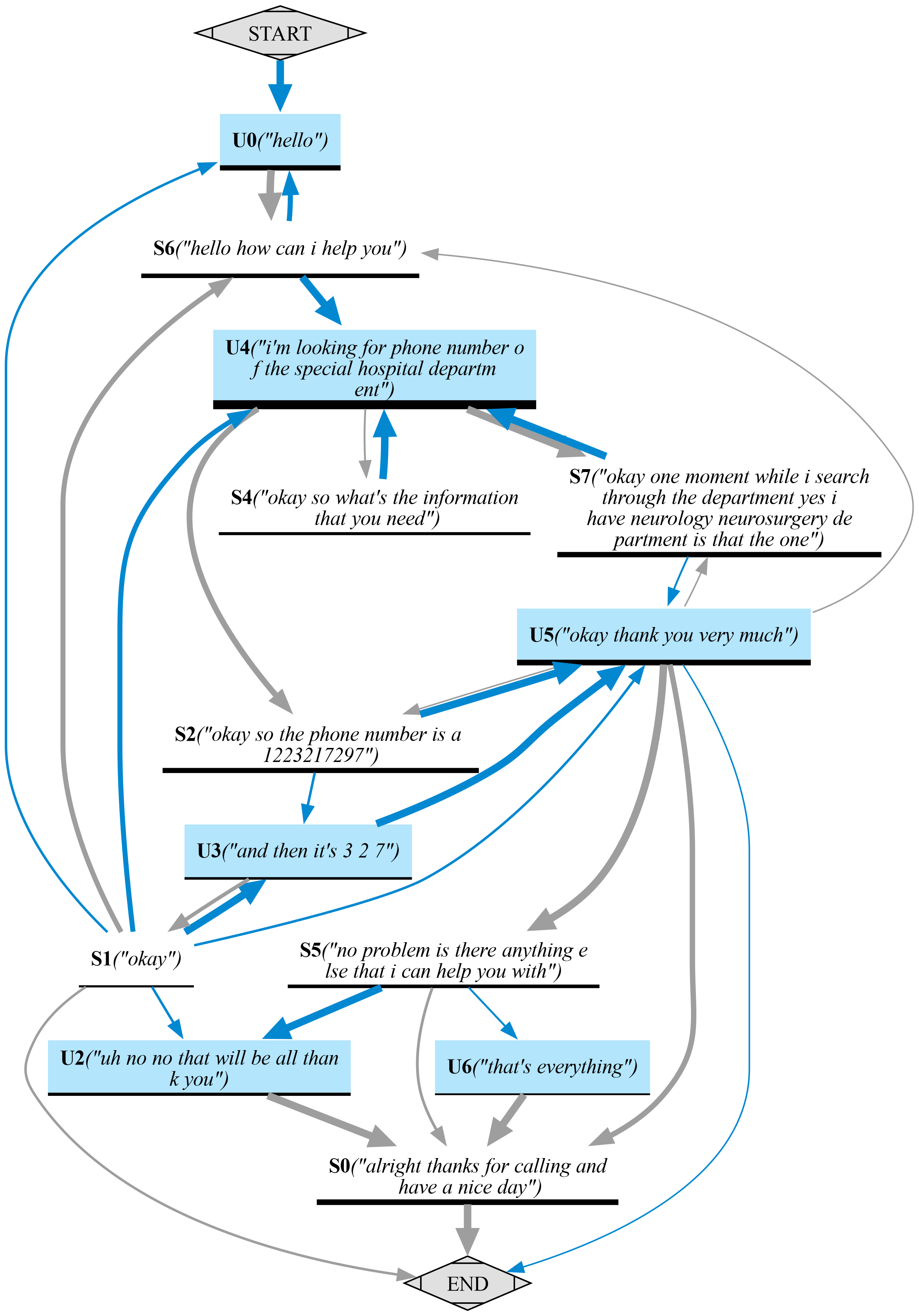}
    \caption{$\hat{G}_{hospital}$ graph obtained with D2F$_{joint}$ containing only one node less than the reference graph in Figure~\ref{fig:gt-graph}.
    Node labels correspond to the cluster ID along a representative utterance (the closest to the cluster centroid).
    Although not the exact same graph as the reference, this graph still allows us to understand the common flow of the conversations with a similar degree of detail:
    first, the user and system greet each other (\textbf{U0} and \textbf{S6}), then the user inform the reason of the call requesting the phone number of a department (\textbf{U4}), the agent may confirm the department (\textbf{S7}) or request more information (\textbf{S4}) before providing the phone number (\textbf{S2}).
    The user may then either confirm the number (\textbf{U3}) or thank the system (\textbf{U5}).
    Finally, the system asks if anything else is required (\textbf{S5}), to which the user may either finish the conversation (\textbf{U6}) or, more likely, thank the system (\textbf{U2}) before the system says goodbye (\textbf{S0}).}
    \label{fig:ours-graph}
\end{figure}

Among the baseline embeddings, DSE stands out (27.90\% average difference across domains), suggesting that conversational-context embeddings are better at capturing the communicative and informative functions of dialog utterances than semantically meaningful embeddings.
Notably, D2F embeddings trained with the proposed soft contrastive loss extract graphs closest in complexity to the references across domains (6.86\% and 8.57\% average difference for D2F$_{single}$ and D2F$_{joint}$, respectively) compared to both D2F-Hard embeddings trained with the vanilla supervised contrastive loss and the other embeddings.
For instance, Figure~\ref{fig:ours-graph} shows the corresponding $\hat{G}_{hospital}$ obtained with D2F$_{joint}$.\footnote{Source code is provided to generate graphs for any given dialogue collection and any embedding model, allowing manual assessment of superior D2F graph quality.}
Finally, it is also worth noting that the D2F graphs are relatively consistent across different domains, even though some domains had only a small amount of in-domain data during training. For instance, the \texttt{hospital} and \texttt{police} domains make up only 0.11\% and 0.07\% of the training set (details in Table~\ref{tab:domains}).

\section{Conclusions}

This paper introduced Dialog2Flow (D2F), embeddings pre-trained for dialog flow extraction grouping utterances by their communicative and informative functions in a latent space.
D2F embeddings were trained on a comprehensive dataset of twenty task-oriented dialog datasets with standardized action annotations, released along with this work.

Future work will enhance D2F embeddings by exploring larger backbone models and advanced methods for sentence embeddings~\cite{jiang2023scaling,jiang-etal-2022-promptbert}. We will also investigate more sophisticated techniques for extracting and representing dialog flows, such as using subtask graphs~\cite{sohn-etal-2023-tod} or adapting dependency parsing for complex dialog structures~\cite{qiu-etal-2020-structured}. Additionally, potential applications include using D2F embeddings to ground LLMs in domain-specific flows for improved transparency and controllability~\cite{raghu-etal-2021-end}, and integrating D2F embeddings into various TOD downstream tasks like dialog state tracking and policy learning.


\section{Limitations}

Our work represents a preliminary exploration with a focus on task-oriented dialogues (TODs) using a relatively simple encoder model. While this work aims to draw attention to this underexplored area, there are a number of limitations that must be acknowledged:

1. \textbf{Scope of Dialogues:} Our study is restricted to task-oriented dialogues. Consequently, the findings and methods may not generalize well to more complex and diverse types of dialogues, particularly those of a non-task-oriented nature. Future research should explore these methods in a broader range of dialogue types to assess their generalizability.

2. \textbf{Domain Specificity:} The model has been trained on a specific collection of domains, dialogue acts, and slots. This limits its ability to generalize to unseen domains or dialogues that involve more complex and varied interactions. Expanding the range of training data to include a wider variety of domains and dialogue types is necessary to improve the model's robustness and applicability.

3. \textbf{Model Complexity:} The encoder model used in this work is relatively standard. There is potential for improvement by employing larger and more advanced models to obtained the final sentence embeddings. 

4. \textbf{Data Size:} Despite being the largest dataset with standardized utterance annotations and the largest spoken TOD dataset, the datasets used in this study are limited in size. Larger datasets are necessary to fully explore and validate the proposed methods. We encourage the research community to build upon this work by utilizing more extensive datasets to enhance the reliability and validity of the results. For instance, perhaps named entity tags may be used as slots to expand annotation beyond pure task-oriented dialogues.

5. \textbf{Evaluation Metrics:} The evaluation metrics employed in this study, while standard, may not capture all aspects of performance relevant to real-world applications. Developing and utilizing a broader set of evaluation metrics would provide a more comprehensive assessment of model performance. Specifically for dialogue flow evaluation, since there is not a standard metric yet, we encourage the research community to explore better ways to represent and quantify the quality of dialogue flows.

By highlighting these limitations, we hope to inspire further research that addresses these challenges, leading to more robust and generalizable solutions building on top of this work.

\section{Ethical Considerations}

We are committed to ensuring the ethical use of our research outcomes. To promote transparency and reproducibility, we will release the source code and pre-trained model weights under the MIT license. This allows for wide usage and adaptation while maintaining open-source principles. 

However, to prevent potential license incompatibilities among the various task-oriented dialogue (TOD) datasets we have utilized, we will not release our unified TOD dataset directly. Instead, we will provide a script that can generate the unified dataset introduced in this paper. This approach allows users to select the specific TOD datasets they wish to include, ensuring compliance with individual dataset licenses.

We acknowledge that gender bias present in the original data could be partially encoded in the embeddings. This may manifest as assumptions about the agent's gender, such as the agent being male or female. We advise users to be aware of this potential bias and encourage further research to mitigate such issues. Continuous efforts to audit and address biases in data and models are essential to ensure fair and equitable AI systems.

\section*{Acknowledgments}

This work was supported by EU Horizon 2020 project ELOQUENCE\footnote{\url{https://eloquenceai.eu/}} (grant number 101070558).
Additionally, this work was inspired by insights gained from the 2023 Jelinek Memorial Summer Workshop on Speech and Language Technologies (JSALT)\footnote{\url{https://jsalt2023.univ-lemans.fr}}, which was partially supported by Johns Hopkins University and the EU project ESPERANTO (grant number 101007666). Our participation in JSALT was further supported by the EU Horizon 2020 project HumanE-AI-Net\footnote{\url{https://www.humane-ai.eu/}} (grant number 952026), under the micro project ``Grounded Dialog Models from Observation of Human-to-Human Conversation''.




\bibliography{main}

\appendix
\setcounter{table}{0}
\setcounter{figure}{0}
\renewcommand{\thetable}{A\arabic{table}}
\renewcommand{\thefigure}{A\arabic{figure}}

\newpage

\section{Unified TOD Dataset}
\label{app:dataset}

\begin{table}[!ht]
    \centering
    \small
    \begin{tabular}{l@{}c@{~}c@{~}c@{}c@{}}
        \toprule
        \multicolumn{5}{p{0.95\linewidth}}{\textbf{Dialog acts:} \tiny \texttt{inform(64.66\%) request(12.62\%) offer(6.62\%) inform\_success(3.07\%) good\_bye(2.67\%) agreement(2.45\%) thank\_you(2.25\%) confirm(2.10\%) disagreement(1.60\%) request\_more(1.06\%) request\_alternative(0.90\%) recommendation(0.70\%) inform\_failure(0.64\%) greeting(0.31\%) confirm\_answer(0.18\%) confirm\_question(0.17\%) request\_update(0.02\%) request\_compare(0.01\%)}} \\
        \midrule
        \multicolumn{5}{p{0.95\linewidth}}{\textbf{Domains:} \tiny \texttt{movie(32.98\%) restaurant(13.48\%) hotel(10.15\%) train(4.52\%) flight(4.30\%) event(3.56\%) attraction(3.50\%) service(2.44\%) bus(2.28\%) taxi(2.21\%) rentalcars(2.20\%) travel(2.16\%) music(1.81\%) medium(1.66\%) ridesharing(1.30\%) booking(1.21\%) home(1.01\%) finance(0.79\%) airline(0.69\%) calendar(0.69\%) fastfood(0.68\%) insurance(0.61\%) weather(0.58\%) bank(0.47\%) hkmtr(0.36\%) mlb(0.35\%) ml(0.31\%) food(0.30\%) epl(0.30\%) pizza(0.25\%) coffee(0.24\%) uber(0.24\%) software(0.23\%) auto(0.21\%) nba(0.20\%) product\_defect(0.17\%) shipping\_issue(0.16\%) alarm(0.13\%) order\_issue(0.13\%) messaging(0.13\%) hospital(0.11\%) subscription\_inquiry(0.11\%) account\_access(0.11\%) payment(0.10\%) purchase\_dispute(0.10\%) nfl(0.09\%) chat(0.08\%) police(0.07\%) single\_item\_query(0.06\%) storewide\_query(0.06\%) troubleshoot\_site(0.06\%) manage\_account(0.06\%)}} \\
        \bottomrule
    \end{tabular}
    \caption{Standardized dialog act and domain labels in our unified TOD datasets, ordered by their proportion of utterances.}
    \label{tab:domains}
\end{table}

Our training data is sourced from a diverse range of TOD datasets meticulously curated in DialogStudio~\cite{zhang-etal-2024-dialogstudio}. DialogStudio comprises over 80 dialog datasets, with 30 focusing on task-oriented conversations.
We conducted a comprehensive manual analysis of these 30 TOD datasets to identify those from which we could extract dialog act and/or slot annotations.
From this analysis, we identified 20 datasets that met our criteria, as summarized in Table~\ref{tab:datasets}.
The datasets in DialogStudio are unified under a consistent format while retaining their original information.
However, this format only unifies the access to the conversations \textit{per se}, omitting annotations and components of task-oriented dialogs.
We then manually inspected each dataset to locate and extract the necessary annotations.
This process involved identifying where and how annotations were stored originally in each dataset, extracting dialog act and/or slot annotations for each turn, either explicitly or implicitly by keeping track of the changes in the dialog state annotation from one turn to the next, and standardizing domain names and dialog act labels across datasets.

To standardize dialog act labels, we mapped the 44 unique labels found across datasets to 18 normalized dialog act labels, informed by the semantic meaning described in the original dataset papers (mapping detailed in Table~\ref{tab:da-labels}).
After this process, we unified all datasets under a consistent format, detailed in the next subsection, incorporating per-turn dialog act and slot annotations. The resulting unified TOD dataset comprises 3.4 million utterances annotated with 18 standardized dialog acts, 524 unique slot labels, and 3,982 unique action labels (dialog act + slots). These annotations span across 52 different domains, as detailed in Table~\ref{tab:datasets}.

Our unified TOD dataset is a valuable resource providing a comprehensive and standardized collection of annotated utterances across diverse domains under a common format.

\subsection{Dataset Format}

Our unified dataset standardizes the TOD datasets into the following common JSON format with per-utterance annotations:

\begin{lstlisting}[language=json,numbers=none,basicstyle=\small]
{
 "stats": {"domains": {...},
           "labels": {..}},
 "dialogs": {
  "<DIALOGUE_ID0>": [
   {
    "speaker": <SPEAKER>,
    "text": <RAW_UTTERANCE>,
    "domains": [...],
    "labels": {
     "dialog_acts": {
      "acts" : [...],
      "main_acts" : [...],
      "original_acts" : [...],
     },
     "slots": [...],
     "intents": [...]
    }
   },
   ...
  ],
  "<DIALOGUE_ID1>": [...],
  ...
 }
}
\end{lstlisting}

The JSON structure has two main parts: a \texttt{"stats"} header and a \texttt{"dialogs"} body.
The \texttt{"stats"} field provides statistics about the labels and domains in the dataset.
The \texttt{"dialogs"} field contains dialog IDs, each linked to a list of annotated utterance objects.
Each utterance object includes its speaker, text, domains, and associated labels for dialog acts, slots, and intents.
Dialog act labels contain the original labels (\texttt{"original\_acts"}) as well as their standardized values (\texttt{"acts"}) and parent values (\texttt{"main\_acts"}) as mapped in Table~\ref{tab:da-labels}.

\begin{figure}[!t]
    \centering
    \includegraphics[width=1\linewidth]{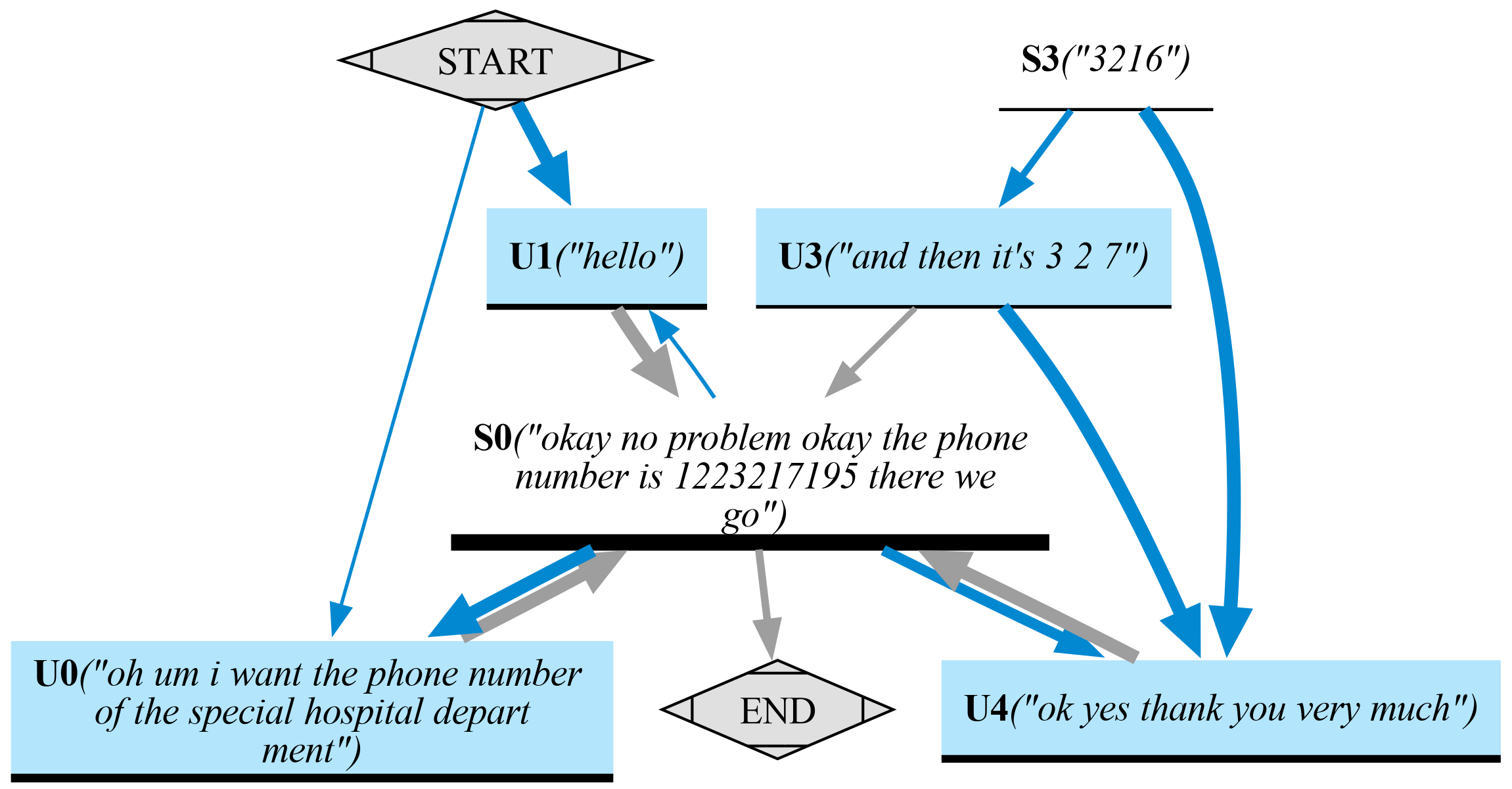}
    \caption{$\hat{G}_{hospital}$ graph obtained with Sentence-BERT (8 nodes/actions in total).
    Node labels correspond to the cluster ID along a representative utterance (the closest to the cluster centroid).}
    \label{fig:sbert-graph}
\end{figure}

\begin{figure}[!t]
    \centering
    \includegraphics[width=1\linewidth]{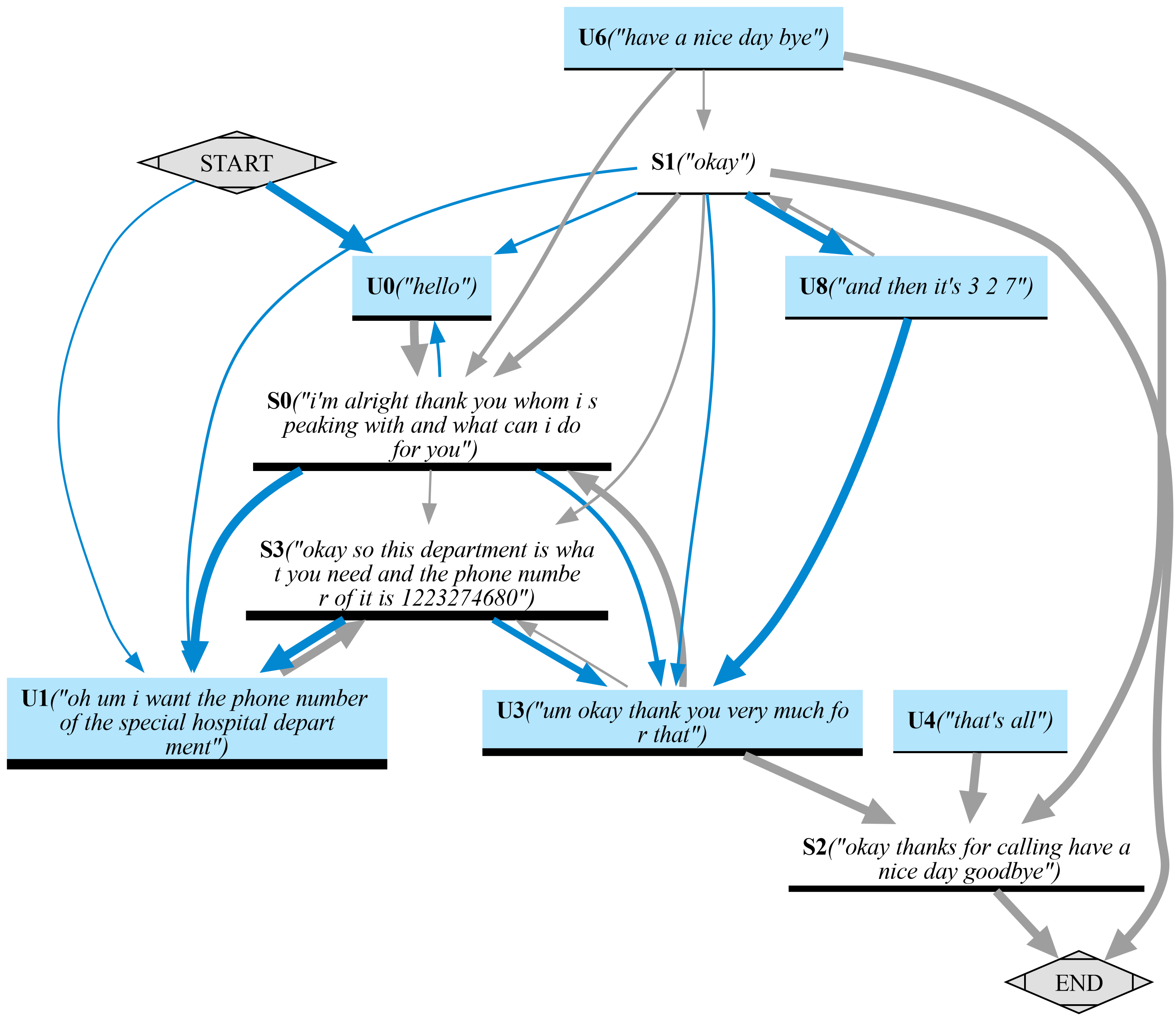}
    \caption{$\hat{G}_{hospital}$ graph obtained with DSE (12 nodes/actions in total).
    Node labels correspond to the cluster ID along a representative utterance (the closest to the cluster centroid).}
    \label{fig:dse-graph}
\end{figure}

\section{Training Details}
\label{app:hyperparameters}

Following the experimental setup of DSE~\cite{zhou-etal-2022-learning} and TOD-BERT~\cite{wu-etal-2020-tod}, we set the contrastive head dimension to $d=128$ and use BERT$_{base}$ as the backbone model for the encoder\footnote{Thus, the embedding size is $n=768$.}. Additional configurations reported in Appendix~\ref{app:ablation}.

For the soft contrastive loss, the semantic similarity measure $\delta(y_i, y_j) = \mathbf{y}_i \cdot \mathbf{y}_j$ was computed using label embeddings $\mathbf{y}$ obtained with the best-performing pre-trained Sentence-BERT model on semantic search, namely the \href{https://huggingface.co/sentence-transformers/multi-qa-mpnet-base-dot-v1}{\texttt{multi-qa-mpnet-base-dot-v1}} model.
As shown in Appendix~\ref{app:ablation}, we also experimented with the \href{https://huggingface.co/sentence-transformers/all-mpnet-base-v2}{\texttt{all-mpnet-base-v2}} model, which has the best average performance among all pre-trained Sentence-BERT models.
The soft label temperature parameter was set to $\tau'=0.35$ after a preliminary study determined it to be a reasonable threshold for both joint and single training targets (Appendix~\ref{app:label_temperature}).

In line with the settings of DSE and TOD-BERT, the learning rates for the contrastive head and the encoder model were set to $3\text{e-}4$ and $3\text{e-}6$, respectively. The contrastive temperature parameter $\tau$ was set to $0.05$. Models were trained for 15 epochs and then saved for evaluation. The maximum sequence length for the Transformer encoder was empirically set to 64 to accommodate at least 99\% of the samples, as most TOD utterances are short.
Finally, the batch size was set to 64 since we found that, contrary to typical self-supervised contrastive learning, larger batch sizes resulted in lower performance.\footnote{A grid search with batch sizes 64, 128, 256, and 512 was performed, training models for one epoch and evaluating the similarity-based 5-shot F$_1$ score on our evaluation set. Larger batch sizes consistently yielded lower scores across all models (both standard and soft supervised contrastive loss models). For instance, DFD$_{joint}$ scored $63.23$, $61.64$, $58.77$, and $56.30$ for batch sizes 64, 128, 256, and 512, respectively.}

\section{Ablation study}
\label{app:ablation}

\begin{table}[!t]
    \centering
    \small
    \begin{tabular}{>{\rowmac}l@{}>{\rowmac}c>{\rowmac}c<{\clearrow}}
        \toprule
        \textbf{DF2 Variation} & \textbf{F$_1$ score} & \textbf{$\Delta$ Anisotropy ($\uparrow$)} \\
        \midrule
        \textbf{D2F-Hard$_{single}$} & 67.82 & 0.332 \\
        \addlinespace
        \cdashline{2-3}
        \addlinespace
        \setrow{\itshape} \hspace{.1in}\textbf{*} DSE Backbone & +2.66 & +0.011  \\
        \setrow{\itshape} \hspace{.1in}\textbf{+} Self-Supervision & -7.41 & -0.002 \\
        
        \midrule
        \textbf{D2F-Hard$_{joint}$} & 66.22 & 0.230 \\
        \addlinespace
        \cdashline{2-3}
        \addlinespace
        \setrow{\itshape} \hspace{.1in}\textbf{*} DSE Backbone & +1.97 & +0.010 \\
        \setrow{\itshape} \hspace{.1in}\textbf{+} Self-Supervision & -6.01 & -0.064 \\

        \midrule
        \textbf{D2F$_{single}$} & 70.89 & 0.597 \\
        \addlinespace
        \cdashline{2-3}
        \addlinespace
        \setrow{\itshape} \hspace{.1in}\textbf{*} DSE Backbone & +0.97 & +0.012 \\
        \setrow{\itshape} \hspace{.1in}\textbf{*} all-mpnet-base-v2 Label & -0.60 & -0.038 \\
        \setrow{\itshape} \hspace{.1in}\textbf{+} Self-Supervision & -6.65 & -0.189 \\
        \setrow{\itshape} \hspace{.1in}\textbf{--} Contrastive Head & -1.13 & -0.047 \\

        \midrule
        \textbf{D2F$_{joint}$} & 70.94 & 0.451 \\
        \addlinespace
        \cdashline{2-3}
        \addlinespace
        \setrow{\itshape} \hspace{.1in}\textbf{*} DSE Backbone & +0.65 & +0.011 \\
        \setrow{\itshape} \hspace{.1in}\textbf{*} all-mpnet-base-v2 Label & -0.34 &  -0.038 \\
        \setrow{\itshape} \hspace{.1in}\textbf{+} Self-Supervision & -8.06 & -0.126 \\
        \setrow{\itshape} \hspace{.1in}\textbf{--} Contrastive Head & -3.78 & -0.073 \\

        \bottomrule
    \end{tabular}
    \caption{Ablation study results for various D2F configurations. Additions, subtractions, and replacements of components are marked with \textbf{+}, \textbf{--}, and \textbf{*} symbols, respectively.
    Values show the impact on 5-shot classification F$_1$ score and anisotropy as reported in Table~\ref{tab:results-in-domain}.}
    \label{tab:ablation}
\end{table}

We conducted an ablation study to evaluate the effects of different configurations on the performance of our D2F models. The following variations were tested:

\begin{itemize}
    \item \textbf{DSE Backbone}: Replacing the original BERT encoder with the pre-trained DSE model.
    \item \textbf{Label Encoder}: Using the Sentence-BERT model \href{https://huggingface.co/sentence-transformers/all-mpnet-base-v2}{\texttt{all-mpnet-base-v2}}, which has the best reported average performance for semantic similarity.
    \item \textbf{Self-Supervision}: Adding the self-supervised loss from DSE ($\mathcal{L}^{self}$) trained jointly with our targets ($\mathcal{L} + \mathcal{L}^{self}$) on the same data as DSE. This was done to evaluate whether jointly training as DSE would yield better performance than using the pre-trained DSE encoder directly as the backbone.
    \item \textbf{Contrastive Head Removal}: Removing the contrastive head used during training.
\end{itemize}

The results of these variations are summarized in Table~\ref{tab:ablation}. The only configuration that consistently improved performance was the replacement of the backbone model with the pre-trained DSE model, increasing the F$_1$ score and anisotropy across all variations.

In contrast, adding self-supervision generally degraded performance, indicating that the additional DSE self-supervised loss $\mathcal{L}^{self}$ may not complement our targets effectively when trained jointly. Similarly, removing the contrastive head during training resulted in a notable performance drop, highlighting its importance.\footnote{Each different configuration required re-training the model for 15 epochs, a process that takes approximately 5 days on a single GeForce RTX 3090 GPU.}

\begin{figure}[t!]
    \centering
    \includegraphics[width=0.5\textwidth]{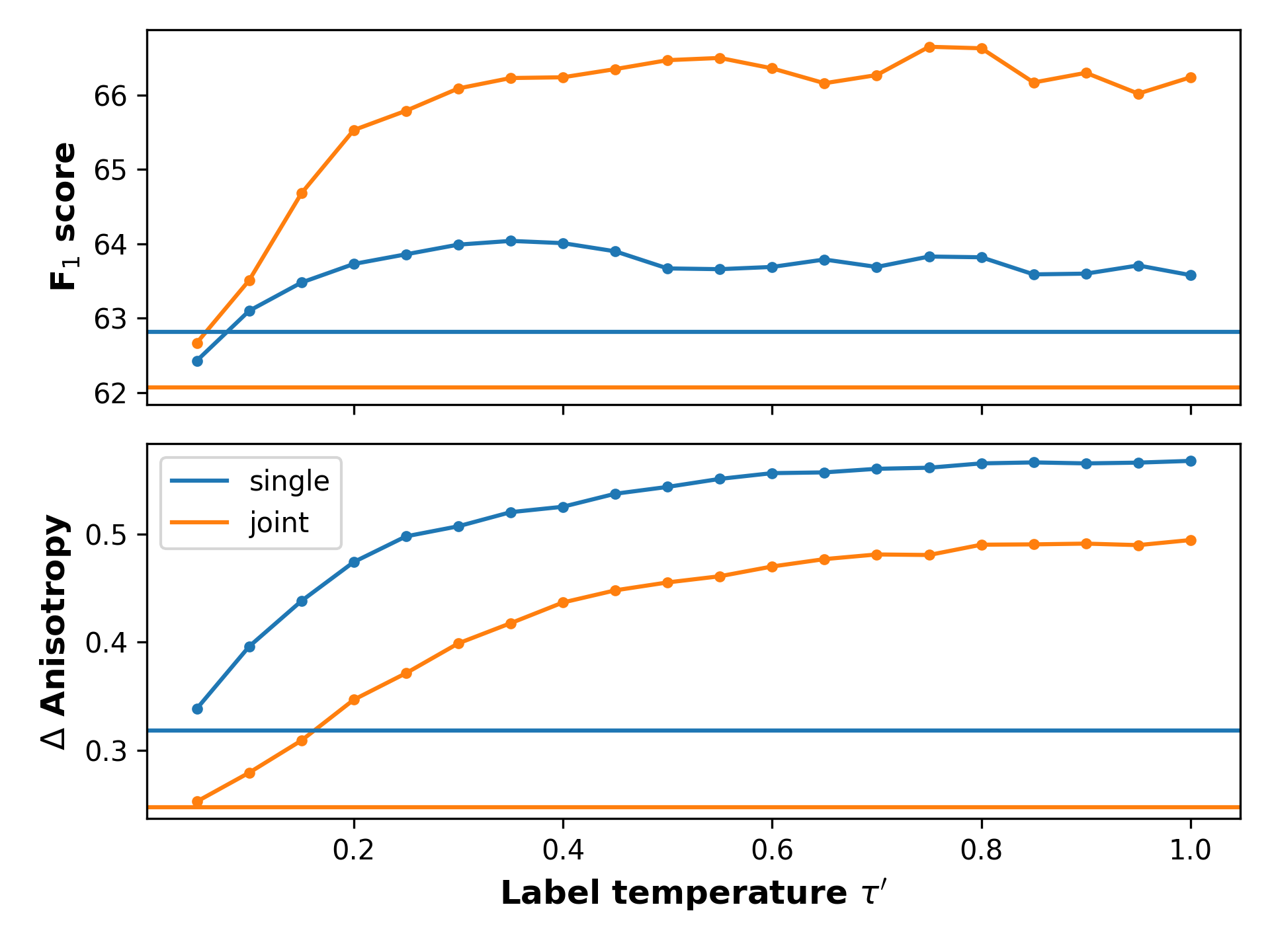}
    \caption{Change in F$_1$ score (top) and $\Delta$ Anisotropy (bottom) with respect to the label temperature $\tau'$ (x-axis). The blue and orange curves represent D2F$_{single}$ and D2F$_{joint}$, respectively. Horizontal lines indicate the performance of their D2F-Hard counterparts using the standard hard supervised contrastive loss.}
    \label{fig:label_temperature}
\end{figure}

\begin{figure*}[t!]
    \centering
    \begin{subfigure}[t]{0.45\textwidth}
        \centering
        \includegraphics[trim={0 1cm 0 1.34cm}, clip, width=\textwidth]{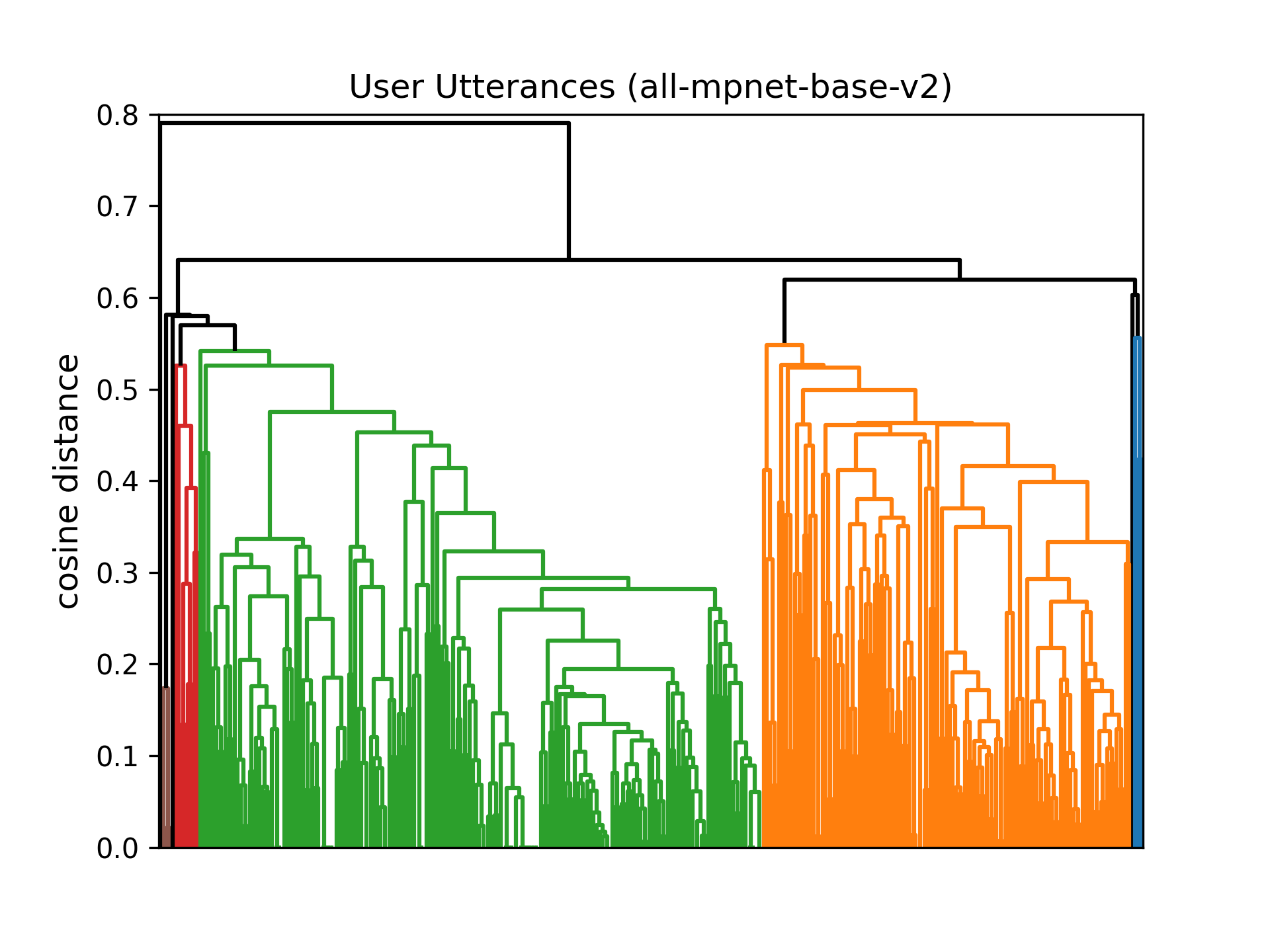}
        \caption{\textbf{Sentence-BERT}}
    \end{subfigure}%
    \begin{subfigure}[t]{0.45\textwidth}
        \centering
        \includegraphics[trim={0 1cm 0 1.34cm}, clip, width=\textwidth]{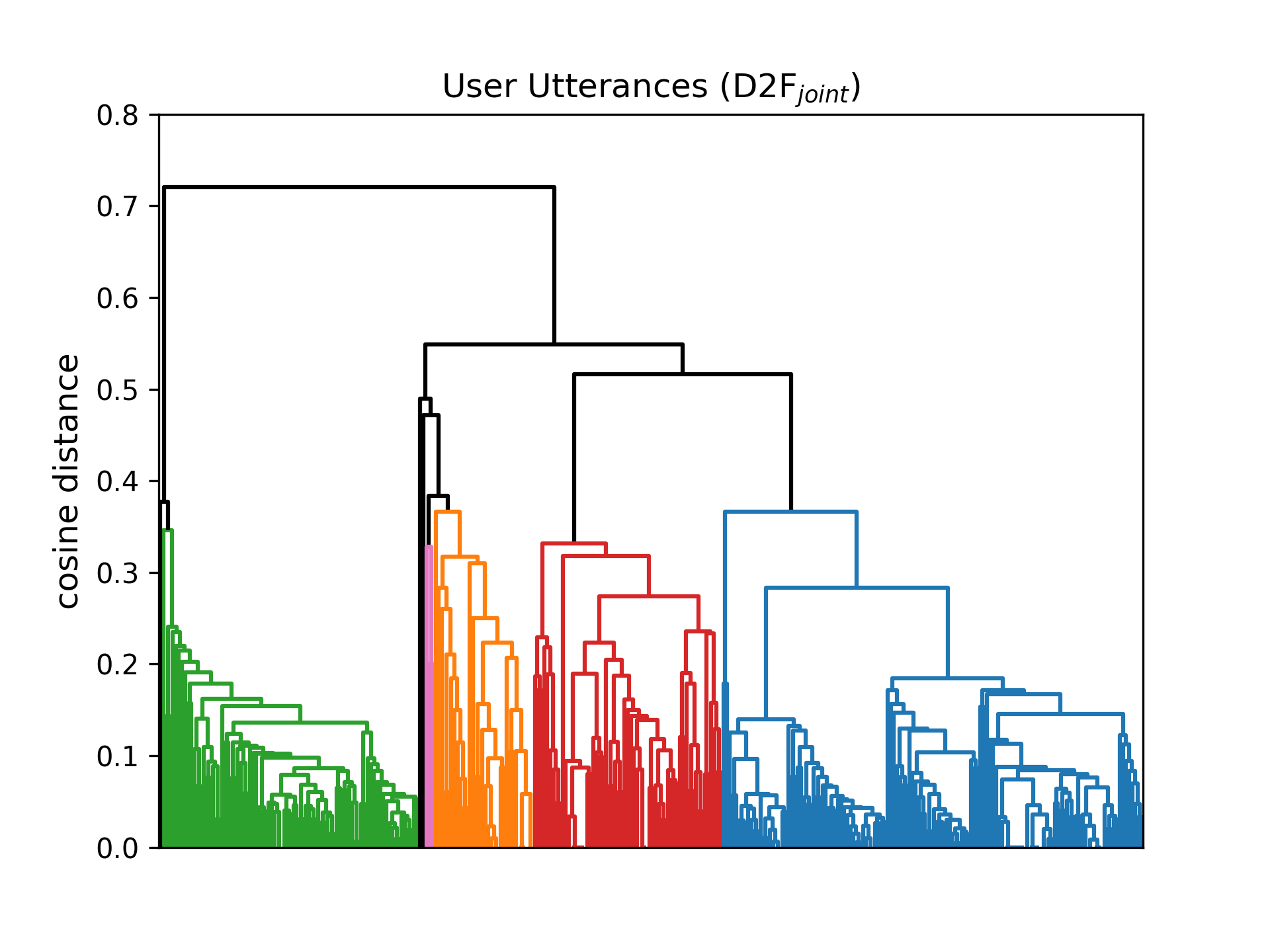}
        \caption{\textbf{D2F$_{joint}$}}
    \end{subfigure}

    \caption{Dendrograms obtained by hierarchically clustering all user utterances in the \texttt{hospital} domain using Sentence-BERT embeddings (left) and D2F$_{joint}$ embeddings (right). The clustering and the plots were obtained using the \texttt{AgglomerativeClustering} class from \texttt{scikit-learn}, with the number of clusters set to 4 (indicated by different colors).}
    \label{fig:dendogram}
\end{figure*}

\section{Supervised Soft Contrastive Loss Explanation}
\label{app:loss}

Let $p(pos\!=\!j\!\mid\!x_i)$ be the probability of $j$-th sample in the batch being positive given the $i$-th anchor. Then, the loss in Equation~\ref{eq:loss} is equivalent to the categorical cross-entropy of correctly classifying the positions in the batch with positive samples for the given $x_i$ anchor:
\begin{equation}
    -\sum_{j=1}^N{p(pos\!=\!j\!\mid\!x_i) \text{log}\ \hat{p}(pos\!=\!j\!\mid\!x_i)}
\label{eq:cross-entropy}
\end{equation}
where the true/target distribution $p$ is defined as
\begin{equation}
p(pos\!=\!j\!\mid\!x_i) = 
     \begin{cases}
       \frac{1}{|\mathcal{P}_i|}, &\quad\text{if}\ y_i = y_j \\
       0, &\quad\text{if}\ y_i \neq y_j \\
     \end{cases}
\label{eq:p}
\end{equation}
and the predicted distribution $\hat{p}$ is an $N$-way softmax-based distribution proportional to the alignment/similarity between (the vectors of) the given $x_i$ anchor and each $x_j^+$ sample:
\begin{align*}
    \hat{p}(pos\!=\!j\!\mid\!x_i) = \frac{e^{\mathbf{z}_i \cdot \mathbf{z}_j^+ / \tau}}{\sum_{k=1}^N{e^{\mathbf{z}_i \cdot \mathbf{z}_k^+ / \tau}}}
\end{align*}
Note that the target distribution in Equation~\ref{eq:p} treats all samples with different labels as equally negative, independently of the semantics of the labels.
However, we hypothesize that better representations can be obtained by taking advantage of the semantics of the labels to model more nuanced relationships. More precisely, let $\delta(y_i,y_j)$ be a semantic similarity measure between both labels, we define a new target distribution $p(pos\!=\!j\!\mid\!x_i)\propto\delta(y_i,y_j)$ as:
\begin{equation}
    p(pos\!=\!j\!\mid\!x_i) = \frac{e^{\delta(y_i, y_j) / \tau'}}{\sum_{k=1}^N{e^{\delta(y_i, y_k) / \tau'}}}
\label{eq:p-new}
\end{equation}
\noindent
where $\tau'$ is the temperature parameter to control how soft/hard the negative labels are (Appendix~\ref{app:label_temperature}).\footnote{On both extremes, sufficiently small $\tau'$ will resemble the original distribution in Equation~\ref{eq:p} while sufficiently large $\tau'$ will resemble a uniform distribution leading to no contrast between positive and negative samples.}
Note that unlike Equation~\ref{eq:p},\footnote{Equation~\ref{eq:p} encourages the encoder to separate all negatives $180^\circ$ away from their anchors: if $y_i \neq y_j$, $\hat{p}(pos\!=\!j\!\mid\!x_i)\to0 \Rightarrow e^{(\cdot)} \to 0 \Rightarrow \mathbf{z}_i \cdot \mathbf{z}_j^+ \to -1$.} this equation allows searching for an encoder that tries to separate anchors and negatives by \textit{degrees proportional to how semantically similar their labels are}.
Therefore, by replacing Equation~\ref{eq:p-new} in Equation~\ref{eq:cross-entropy}, our soft contrastive loss is finally defined as:

\begin{align*}
    \ell_i^{soft}\!=\!-\sum_{j=1}^N \frac{e^{\delta(y_i, y_j) / \tau'}}{\sum_{k=1}^N{e^{\frac{\delta(y_i, y_k)}{\tau'}}}} \text{log} \frac{e^{\mathbf{z}_i \cdot \mathbf{z}_j^+ / \tau}}{\sum_{k=1}^N{e^{\frac{\mathbf{z}_i \cdot \mathbf{z}_k^+}{\tau}}}}
\end{align*}

\section{Soft Contrastive Loss Temperature}
\label{app:label_temperature}

To understand the benefits of the "softness" introduced by our proposed contrastive loss compared to the conventional hard supervised contrastive loss, we conducted a preliminary study examining the impact of the label temperature parameter $\tau'$. We trained models over three epochs, varying the temperature $\tau'$ across a range of values from $0.05$ to $1.0$ in increments of $0.05$. This resulted in 42 different model variants: 20 each for D2F$_{single}$ and D2F$_{joint}$, and one for each D2F-Hard counterpart.

For each $\tau'$ value, we recorded the 5-shot classification F$_1$ score and $\Delta$ anisotropy values as outlined in Section~\ref{sec:evaluation-similarity}. The results are depicted in Figure~\ref{fig:label_temperature}. 

The plots reveal that as the temperature $\tau'$ increases from 0, indicating a transition from hard to softer negative labels, both F$_1$ scores and $\Delta$ anisotropy values improve beyond those obtained with the standard supervised contrastive loss. For both D2F$_{single}$ and D2F$_{joint}$ models, increasing the temperature leads to greater separation between intra-class and inter-class embeddings, as indicated by higher $\Delta$ anisotropy values.

The performance metrics exhibit a steady rise up to a temperature around between $0.35$ and $0.4$, beyond which $\Delta$ anisotropy values begin to plateau and F$_1$ scores become less stable. The advantage of using softer contrast is more pronounced for the joint target (D2F$_{joint}$, represented by the orange line), as evidenced by the larger gap between the orange curve and its corresponding horizontal line (D2F-Hard$_{joint}$).

However, it's important to note that these improvements diminish with additional training epochs. The final difference in performance metrics between soft and hard labels narrows after extended training, as reflected in the results reported in Table~\ref{tab:results-in-domain}, where models were trained for 15 epochs.

\section{How Many Actions to Cluster?}
\label{app:cluster}

In practice, determining the optimal number of clusters (actions) in dialog flow extraction is challenging because it directly affects the granularity of the extracted flows. Hierarchical clustering algorithms, such as agglomerative clustering, are preferred over centroid-based methods like k-means because they provide a visual representation of the data's hierarchical structure, which can be examined to decide the number of clusters or set a distance threshold.

Figure~\ref{fig:dendogram} illustrates dendrograms obtained by hierarchically clustering user utterances in the \texttt{hospital} domain using Sentence-BERT embeddings and D2F$_{joint}$ embeddings. The clustering and plotting were performed using the \texttt{AgglomerativeClustering} class from \texttt{scikit-learn}, with the number of clusters set to 4, represented by different colors.

The dendrograms reveal notable differences between the embeddings. The Sentence-BERT dendrogram (left) shows a structure with two main (semantic) groups with low variability in the distances between child and parent nodes, resulting in a more stretched plot. In contrast, the D2F$_{joint}$ dendrogram (right) displays a clearer separation into four main groups, with larger gaps between child and parent nodes at a certain level of the hierarchy, indicating distinct clusters.
D2F$_{joint}$ embeddings were trained to minimize intra-action distances (pushing them towards the bottom of the dendrogram) and maximize inter-action distances (pushing parent nodes towards the top) facilitating easier identification of clusters. For instance, in the D2F$_{joint}$ dendrogram, the number of actions could be estimated to be between 4 and 7, or a distance threshold around 0.4 could be used to form the clusters.

In our experiments (Section~\ref{sec:evaluation-similarity}), we used the ground truth number of clusters from annotations to ensure consistency in evaluation across the different embeddings.
However, agglomerative clustering was employed to mimic closer a realistic scenario where the number of actions is not predefined.

Thus, hierarchical clustering methods provide a practical approach for approximating the number of actions in practice when such number is unknown.

\section{Deriving Action Labels from Clusters}
\label{app:labels}

\begin{figure}[!ht]
    \centering
    \includegraphics[width=1.1\linewidth]{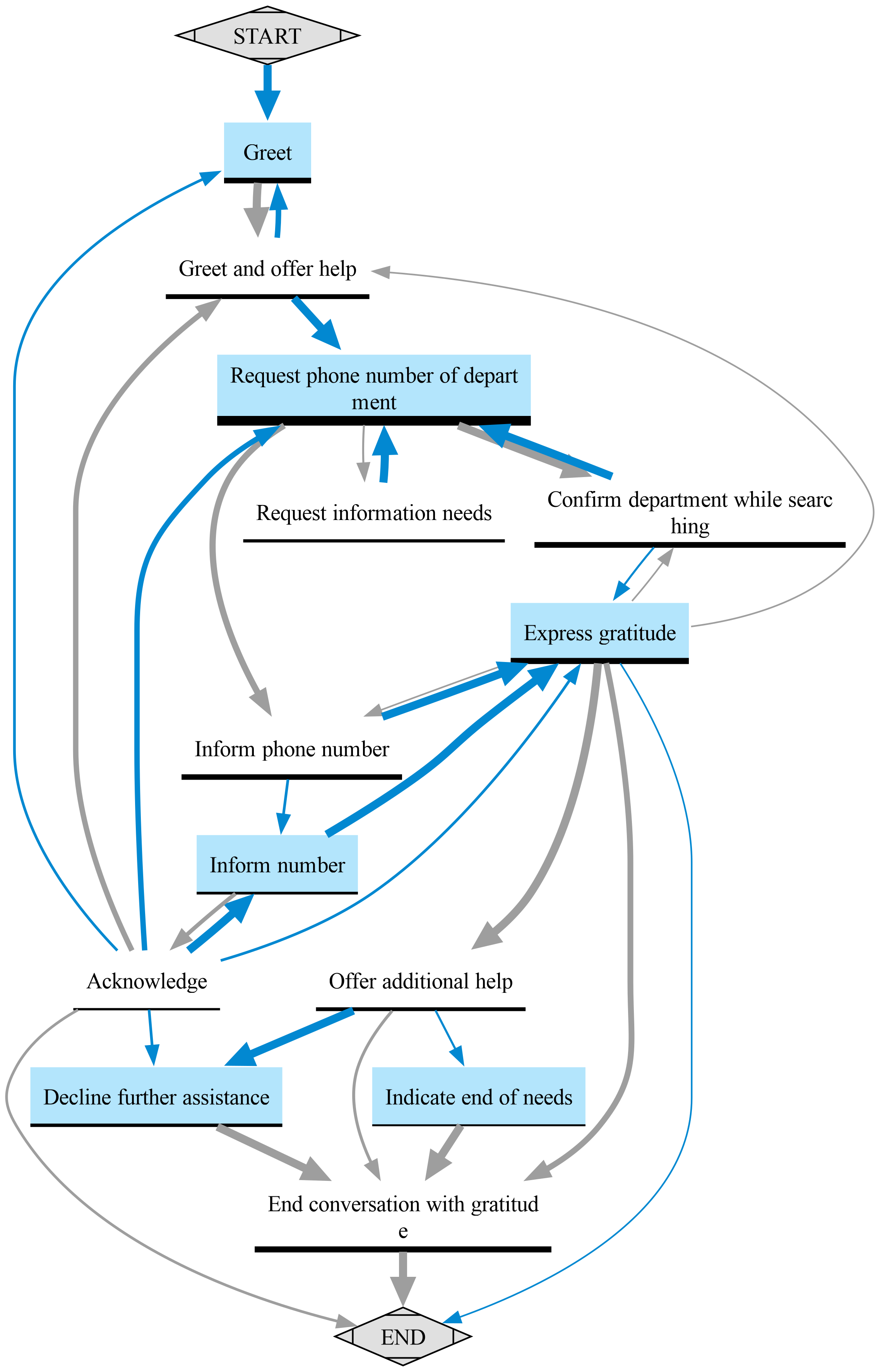}
    \caption{$\hat{G}_{hospital}$ graph from Figure~\ref{fig:ours-graph} with cluster labels generated with ChatGPT.}
    \label{fig:ours-graph-labels}
\end{figure}

In practice, as illustrated in Figures~\ref{fig:sbert-graph}, \ref{fig:dse-graph}, and \ref{fig:ours-graph}, actions are identified by cluster IDs after clustering. However, for certain tasks, such as manual analysis of the extracted dialogue flow, a descriptive action name representing the cluster may be necessary. Following a prompt-based approach similar to that of \citet{sreedhar2024unsupervised} for creating weak intent labels, we can leverage instruction-tuned LLMs to assign representative labels to each cluster based on its constituent utterances. For instance, using the latest OpenAI GPT-4 model (\texttt{gpt-4o}) with the following prompt, where "\texttt{<CLUSTER\_UTTERANCES>}" is replaced with the utterances of a given cluster:

\begin{lstlisting}[language=json,numbers=none,basicstyle=\tiny]
[{"role": "system",
 "content": """Your task is to annotate conversational utterances with the intent expressed as canonical forms. A canonical form is a short summary representing the intent of a set of utterances - it is neither too verbose nor too short.
Be aware that required canonical forms should avoid containing specific names or quantities, only represent the intent in abstract terms.
For example, for:

For the following utterances:
    1. Uh yes i'm looking for a place for entertainment that is in the center of the city
    2. i would like to know where a place for entertainment that is not far away from my location
Canonical form is: "request entertainment place and inform location"

For the following utterances:
    1. Okay so the phone number is a 1223217297
    2. Sure, my phone number is four four five five
    3. 2 3 4 5 6 is her phone number
Canonical form is: "inform phone number"

For the following utterances:
    1. 8 4 0
    2. yes five five three
Canonical form is: "inform number"
"""},
{"role": "user", "content": """Give the following list of utterance provide a single canonical name that represent all of them:
<CLUSTER_UTTERANCES}>"""},
{"role": "assistant", "content": 'The canonical name that represent the above utterances is: "'}]
\end{lstlisting}

Replacing the node labels in Figure~\ref{fig:ours-graph} with those generated by the above prompt yields a more interpretable version of the graph, as shown in Figure~\ref{fig:ours-graph-labels}.

\begin{table*}[!htbp]
    \centering
    \small
    \begin{tabular}{l|l|l}
    \toprule
    \textbf{Original} & \textbf{Standardized}  & \textbf{Parent} \\
    \midrule
    \texttt{inform}               & \texttt{\textbf{inform}} (slots)       & \multirow{12}{*}{\texttt{inform}}         \\
    \addlinespace
    \cdashline{2-2}
    \texttt{notify\_fail}         & \multirow{9}{*}{\textbf{\texttt{inform\_failure}}}      &                \\
    \texttt{notify\_failure}      &                      &                \\
    \texttt{no\_result}           &                      &                \\
    \texttt{nobook}               &                      &                \\
    \texttt{nooffer}              &                      &                \\
    \texttt{sorry}                &                      &                \\
    \texttt{cant\_understand}     &                      &                \\
    \texttt{canthelp}             &                      &                \\
    \texttt{reject}               &                      &                \\
    \addlinespace
    \cdashline{2-2}
    \texttt{book}                 & \multirow{3}{*}{\textbf{\texttt{inform\_success}}}      &                \\
    \texttt{offerbooked}          &                      &                \\
    \texttt{notify\_success}      &                      &                \\
    \addlinespace
    \cdashline{2-3}
    \texttt{request}              & \textbf{\texttt{request}} (slots)      & \multirow{9}{*}{\texttt{request}}        \\
    \texttt{request\_alt}         & \textbf{\texttt{request\_alternative}} &                \\
    \texttt{request\_compare}     & \textbf{\texttt{request\_compare}}     &                \\
    \texttt{request\_update}      & \textbf{\texttt{request\_update}}      &                \\
    \addlinespace
    \cdashline{2-2}
    \texttt{req\_more}            & \multirow{4}{*}{\textbf{\texttt{request\_more}}}        &                \\
    \texttt{request\_more}        &                      &                \\
    \texttt{moreinfo}             &                      &                \\
    \texttt{hearmore}             &                      &                \\
    \addlinespace
    \cdashline{2-3}
    \texttt{confirm}              & \textbf{\texttt{confirm}} (slots)      & \multirow{3}{*}{\texttt{confirmation}}        \\
    \texttt{confirm\_answer}      & \textbf{\texttt{confirm\_answer}}      &                \\
    \texttt{confirm\_question}    & \textbf{\texttt{confirm\_question}}    &                \\
    \addlinespace
    \cdashline{2-3}
    \texttt{affirm}               & \multirow{2}{*}{\textbf{\texttt{agreement}}}            & \multirow{2}{*}{\texttt{agreement}}      \\
    \texttt{affirm\_intent}       &                      &                \\
    \addlinespace
    \cdashline{2-3}
    \texttt{negate}               & \multirow{3}{*}{\textbf{\texttt{disagreement}}}         & \multirow{3}{*}{\texttt{disagreement}}   \\
    \texttt{negate\_intent}       &                      &                \\
    \texttt{deny}                 &                      &                \\
    \addlinespace
    \cdashline{2-3}
    \texttt{offer}                & \multirow{4}{*}{\textbf{\texttt{offer}}}                & \multirow{4}{*}{\texttt{offer}}          \\
    \texttt{select}               &                      &                \\
    \texttt{multiple\_choice}     &                      &                \\
    \texttt{offerbook}            &                      &                \\
    \addlinespace
    \cdashline{2-3}
    \texttt{suggest}              & \multirow{2}{*}{\textbf{\texttt{recommendation}}}       & \multirow{2}{*}{\texttt{recommendation}} \\
    \texttt{recommend}            &                      &                \\
    \addlinespace
    \cdashline{2-3}
    \texttt{greeting}             & \multirow{2}{*}{\textbf{\texttt{greeting}}}             & \multirow{2}{*}{\texttt{greeting}}       \\
    \texttt{welcome}              &                      &                \\
    \addlinespace
    \cdashline{2-3}
    \texttt{thank\_you}           & \multirow{3}{*}{\textbf{\texttt{thank\_you}}}           & \multirow{3}{*}{\texttt{thank\_you}}     \\
    \texttt{thanks}               &                      &                \\
    \texttt{thankyou}             &                      &                \\
    \addlinespace
    \cdashline{2-3}
    \texttt{good\_bye}            & \multirow{3}{*}{\textbf{\texttt{good\_bye}}}            & \multirow{3}{*}{\texttt{good\_bye}}       \\
    \texttt{goodbye}              &                      &                \\
    \texttt{closing}              &                      &                \\
    \bottomrule
    \end{tabular}
    \caption{The original 44 dialog acts with their respective 18 standardized names used to unify all the datasets, along with a parent category grouping them further into 10 parent acts.}
    \label{tab:da-labels}
\end{table*}

\end{document}